\documentclass{article}

\usepackage[preprint]{neurips_2019}

\usepackage{import}
\usepackage{multirow}
\usepackage{graphicx}
\usepackage{float}
\usepackage{wrapfig}
\usepackage{subfig}
\usepackage{tcolorbox}

\usepackage[utf8]{inputenc} 
\usepackage[T1]{fontenc}    
\usepackage{hyperref}       
\usepackage{url}            
\usepackage{booktabs}       
\usepackage{amsfonts}       
\usepackage{amsmath}
\usepackage{nicefrac}       
\usepackage{microtype}      

\graphicspath{{./Fig/}}

\title{Complex-valued neural networks for machine learning on non-stationary physical data}

\author{
  Jesper S\"oren Dramsch\thanks{Corresponding author.} \\
  Technical University of Denmark\\
  Building 375, 2800 Kongens Lyngby \\
  Denmark\\
  \texttt{jesper@dramsch.net}
   \And
  Mikael L\"uthje \\
  Technical University of Denmark\\
  Building 375, 2800 Kongens Lyngby \\
  Denmark\\
  \texttt{mikael@dtu.dk} \\
  \AND
  Anders Nymark Christensen\\
  Technical University of Denmark\\
  Building 324, 2800 Kongens Lyngby \\
  Denmark\\
  \texttt{anym@dtu.dk} \\
}

\begin{document}
\maketitle

\begin{tcolorbox}[title=Pre-Review Preprint from \href{https://arxiv.org/abs/1905.12321}{arXiv:1905.12321}]
    \centering 
    This is manuscript is a preprint and has been submitted for publication. Please note that, the manuscript has yet to be undergo peer-review. Subsequent versions of this manuscript may have different content. Please feel free to contact the corresponding author.\\\textbf{We welcome feedback.}
\end{tcolorbox}

\begin{abstract}
Deep learning has become an area of interest in most scientific areas, including physical sciences. Modern networks apply real-valued transformations on the data. Particularly, convolutions in convolutional neural networks discard phase information entirely. Many deterministic signals, such as seismic data or electrical signals, contain significant information in the phase of the signal. We explore complex-valued deep convolutional networks to leverage non-linear feature maps. Seismic data commonly has a lowcut filter applied, to attenuate noise from ocean waves and similar long wavelength contributions. Discarding the phase information leads to low-frequency aliasing analogous to the Nyquist-Shannon theorem for high frequencies. In non-stationary data, the phase content can stabilize training and improve the generalizability of neural networks. While it has been shown that phase content can be restored in deep neural networks, we show how including phase information in feature maps improves both training and inference from deterministic physical data. Furthermore, we show that the reduction of parameters in a complex network outperforms larger real-valued networks.

\end{abstract}


\section{Introduction}
Seismic data is high-dimensional physical data.
During acquisition, the data is collected over an area on the Earth’s surface.
This images a 3D cube of the subsurface.
Due to low reflection coefficients and low signal-to-noise ratio, the measurements are repeated, while moving over the target area.
This provides a collection of illumination angles over a subsurface area.
The dimensionality of this data has historically been reduced to a stacked 3D cube or 2D sections for interpreters to be able to grasp the information of the seismic data.

With the recent revolution of image classification, segmentation and object detection through deep learning \citep{krizhevsky2012imagenet}, geophysics has regained interest in automatic seismic interpretation (classification), and analysis of seismic signals.
Through transfer learning, several initial successes were presented in \citet{dramsch2018deep}.
Nevertheless, seismic data has its caveats due to the complicated nature of bandwidth-limited wave-based imaging.
Common problems are cycle-skipping of wavelets and nullspaces in inversion problems \citep{yilmaz2001seismic}.

Automatic seismic interpretation is complicated, as the modelling of seismic data is computationally expensive and often proprietary.
Seismic field data is often not available and their interpretation is highly subjective and ground truth is not available.
The lack of training data has been delaying the adoption of existing methods and hindering the development of specific geophysical deep learning methods.
Incorporating domain knowledge into general deep learning models has been successful in other fields \citep{paganini_calogan}. 

The state-of-the-art method has been a iterative windowed Fourier transform for phase reconstruction \citep{Griffin}. 
Modern neural audio synthesis focuses on methods that do not require explicit reconstruction of the phase \citep{Mehri2016, wavenet, wavenet2, waveglow}.
\citet{Mehri2016} introduced a recurrent neural network formulation, where \citet{wavenet} reformulated the synthesis network in a strided convolutional  network.
The original WaveNet formulation in \citet{wavenet} is slow due to the autoregressive filter, warranting the parallel formulation in \citet{wavenet2}.

We explicitly incorporate phase information in a deep convolutional neural network.
These have been heavily explored in the digital signal processing community, before the recent renaissance of neural networks and deep learning.
Relevant examples to seismic data processing include source separation \citep{Scarpiniti2008}, adaptive noise reduction \citep{Suksmono2002}, and optical flow \citep{Miyauchi} with complex-valued neural networks.
\citet{Saroff2018} gives a comprehensive overview of applications of complex-valued neural networks in signal and image processing.

In this work, we calculate the complex-valued seismic trace by applying the Hilbert transform to each trace.
Phase information has been shown to be valuable in the processing \citep{Liner2002} and interpretation of seismic data \citep{Roden1999,Mavko2003}.
\citet{Purves2014} provides a tutorial that shows the implementation details of Hilbert transforms.

In this paper we give a brief overview of convolutional neural networks and then introduce the extension to complex neural networks and seismic data.
We show that including explicit phase information provides superior results to real-valued convolutional neural networks for seismic data.
Difficult areas that contain seismic discontinuities due to geologic faulting are resolved better without leakage of seismic horizons.
We train and evaluate several complex-valued and real-valued auto-encoders to show and compare these properties.
These results can be directly extended to automatic seismic interpretation problems.
\section{Complex Convolutional Neural Networks}
\subsection{Basic principles}
\begin{figure}[H]%
    \centering
    \vspace*{-.75cm}
    \subfloat[Complex Neural Network\label{fig:cmplx}]{\includegraphics[width=.4\linewidth]{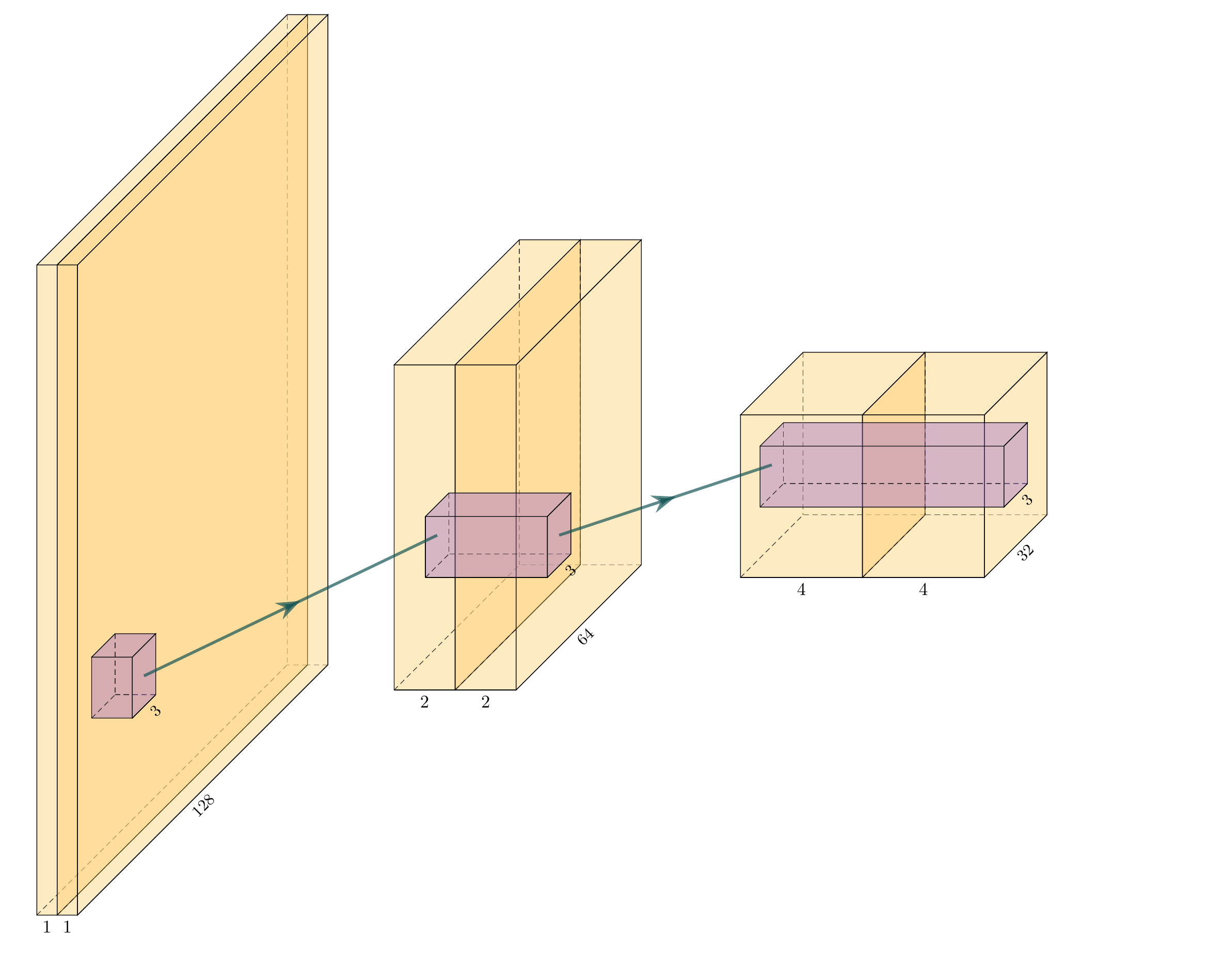}}
    \hfill
    \subfloat[Real Neural Network\label{fig:real}]{\includegraphics[width=.4\linewidth]{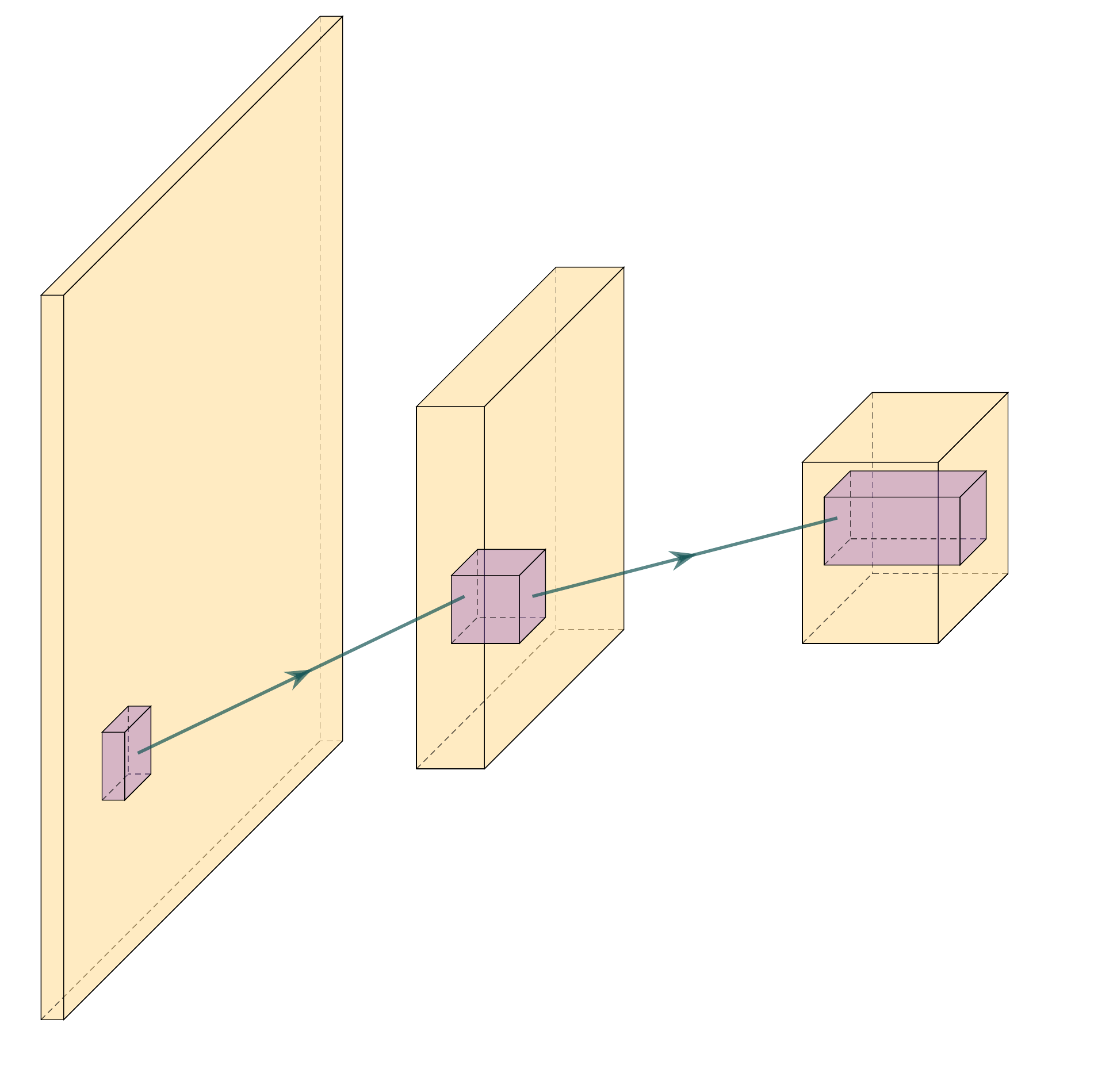}}
    \caption{Schematic of equivalent complex- and real-valued convolutional neural network}%
    \label{fig:3}%
\end{figure}

Convolutional neural networks \citep{lecun1999object} use multiple layers of convolution and subsampling to extract relevant information from the data (see Figure~\ref{fig:3})

The input image is repeatedly convolved with filters and subsampled. This creates many, but smaller and smaller images. 
For a classification task, the final step is then a weighting of these very small images leading to a decision about what was in the original image. 
The filters are learned as part of the training process by exposing the network to training images.
The salient point is, that the convolution kernels are learned based on the training. 
If the goal is - for example - to classify geological facies, the convolutional kernels will learn to extract information from the input, that helps with that task. 
It is thus a very strong methodology, that can be adapted to many tasks.

\subsection{Real- and Complex-valued Convolution}
\label{sec:conv}
Convolution is an operation on two signals f and g or a signal and a filter that produce a third signal, containing information from both of the inputs. 
An example is the moving average filter, which smoothes the input, acting as a low-pass filter. 
Convolution is defined as
\begin{equation}
    f(t)*g(t)=\int_{-\infty}^\infty f(\tau)g(t-\tau)d\tau,
\end{equation}
at the location $\tau$. 
While often applied to real value signals, convolution can be used on complex signals. 
For the integral to exist both $f$ and $g$ must decay when approaching infinity.
Convolution is directly generalizable to N-dimensions by multiple integrations and maintains commutativity, distributivity, and associativity.
In digital signals this extends to discrete values by replacing the integration with summation.

\subsection{Complex Convolutional Neural Networks}
\begin{figure}
    \centering
    \includegraphics[width=\linewidth]{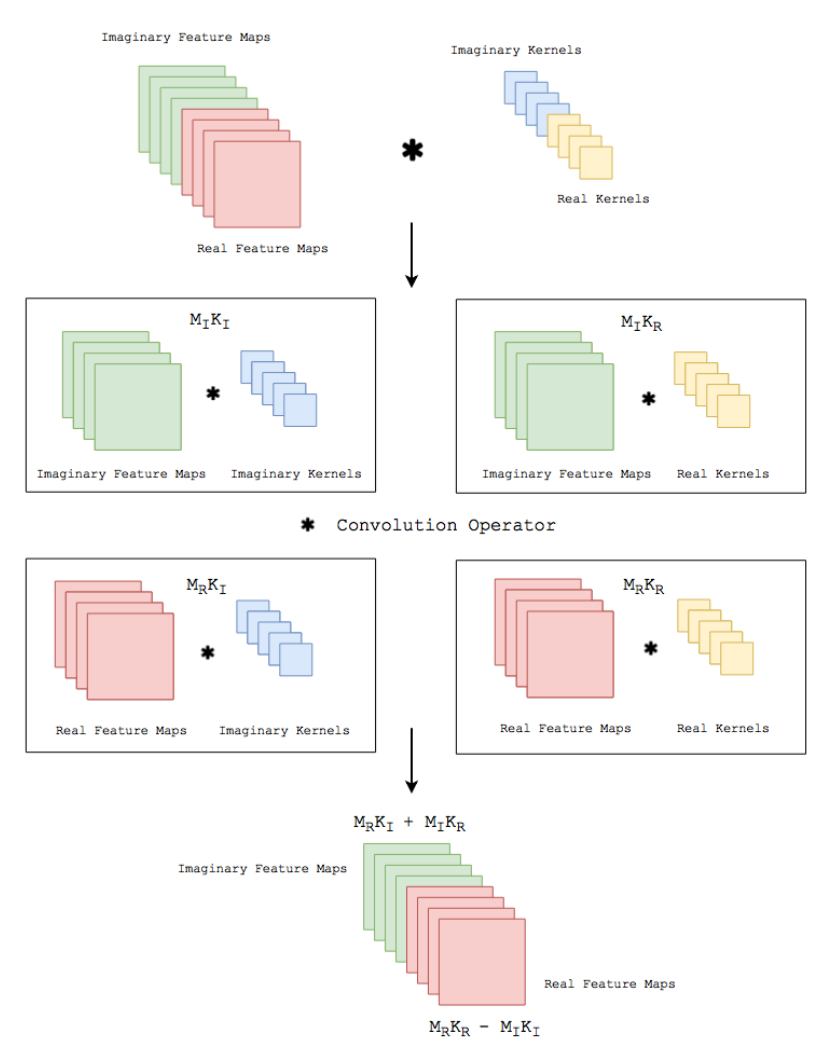}
    \caption{Implementation details of Complex Convolution CC-BY (Trabelski et al. 2017).}
    \label{fig:4}
\end{figure}
Complex convolutional networks provide the benefit of explicitly modelling the phase space of physical systems \citep{trabelsi2017deep}. 
The complex convolution introduced in Section~\ref{sec:conv}, can be explicitly implemented as convolutions of the real and complex components of both kernels and the data.
A complex-valued data matrix in cartesian notation is defined as $\textbf{M} = M_\Re + i M_\Im$ and equally, the complex-valued convolutional kernel is defined as $\textbf{K} = K_\Re + i K_\Im$.
The individual coefficients $(M_\Re, M_\Im, K_\Re, K_\Im)$ are real-valued matrices, considering vectors are special cases of matrices with one of two dimensions being one.

Solving the convolution of 
\begin{equation}
M' = K * M = (M_\Re + i M_\Im) * (K_\Re + i K_\Im),
\end{equation}
we can apply the distributivity of convolutions (cf.~section~\ref{sec:conv}) to obtain
\begin{equation}
M' =  \{M_\Re * K_\Re - M_\Im * K_\Im\} + i \{ M_\Re * K_\Im + M_\Im * K_\Re\},
\end{equation}
where $K$ is the Kernel and $M$ is a data vector (see Figure~\ref{fig:4}).

We can reformulate this in algebraic notation
\begin{equation}
    \begin{bmatrix} \Re\{M * K\} \\ \Im\{M * K\} \end{bmatrix} = \begin{bmatrix} K_{\Re} & -K_{\Im} \\ K_{\Im} & K_{\Re} \end{bmatrix} * \begin{bmatrix}  M_{\Re} \\ M_{\Im} \end{bmatrix}
\end{equation}

Complex convolutional neural networks learn by back-propagation. 
\citet{Sarroff2015} state that the activation functions, as well as the loss function must be complex differentiable (holomorphic).
\citet{trabelsi2017deep} suggest that employing complex losses and activation functions is valid for speed, however, refers that \citet{Hirose2012} show that complex-valued networks can be optimized individually with real-valued loss functions and contain piecewise real-valued activations.
We reimplement the code \citet{trabelsi2017deep} provides in keras \citep{keras} with tensorflow \citep{tensorflow}, which provides convenience functions implementing a multitude of real-valued loss functions and activations.

While common up- and downsampling functions like MaxPooling, UpSampling, or striding do not suffer from complex-valued neural networks, batch normalization (BN) \citep{ioffe2015batch} does.
Real-valued batch normalization normalizes the data to zero mean and a standard deviation of 1. 
This does not guarantee normalization in complex values. 
\citet{trabelsi2017deep} suggest implementing a 2D whitening operation as normalization in the following way.
\begin{equation}
\widetilde{x} = V^{-\frac{1}{2}} ( x - \mathbb{E}[x] ),    
\end{equation}
where $x$ is the data and $V$ is the 2x2 covariance matrix, with the covariance matrix being
\begin{equation}
    V = \begin{bmatrix} V_{\Re\Re} & V_{\Re\Im} \\ V_{\Im\Re} & V_{\Im\Im} \end{bmatrix}
\end{equation}
Effectively, this multiplies the inverse of the square root of the covariance matrix with the zero-centred data. This scales the covariance of the components instead of the variance of the data \citep{trabelsi2017deep}.
\subsection{Auto-encoders}
\begin{figure}
    \centering
    \includegraphics[width=\linewidth]{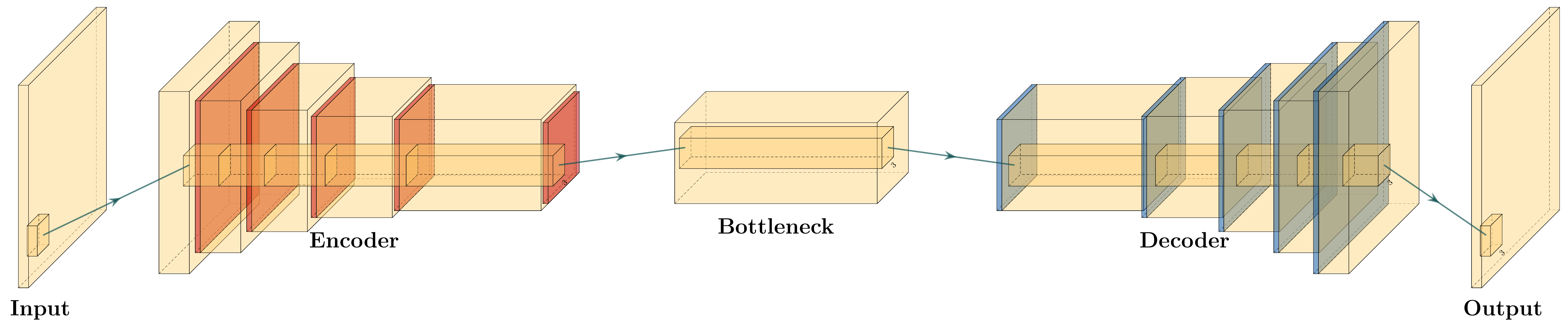}
    \caption{Typical autoencoder architecture. The data is compressed to a low dimensional bottleneck, and then reconstructed.}
    \label{fig:5}
\end{figure}

Auto-encoders \citep{hinton2006reducing} are a special configuration of the encoder-decoder network that map data to a low-level representation and back to the original data. 
This low-level representation is often called bottleneck or code layer.
Auto-encoder networks map $f(x) = x$, where $x$ is the data and $f$ is an arbitrary network. 
The architecture of auto-encoders is an example of lossy compression and recovery from the lossy representation. Commonly, recovered data is blurred by this process.

The principle is illustrated in figure~\ref{fig:5}. 
The input is transformed to a low-dimensional representation - called a code or latent space - and then reconstructed again from this low dimensional representation. 
The intuition is, that the network has to extract the most salient parts from the data, to be able to perform a reconstruction. 
As opposed to other methods for dimensionality reduction - e.g. principal component analysis - an auto-encoder can find a non-linear representation of the data. 
The low-dimensional representation can then be used for anomaly detection, or classification.
\section{Aliasing in Patch-based training}
\subsection{Mean-Shift in Neural Networks}
A single neuron in a neural network can be described by $\sigma ( w \cdot x + b )$, where $w$ is the network weights, $x$ is the input data, $b$ is the network bias, and $\sigma$ is a non-linear activation function. During training, the network weights $w$ and biases $b$ are are adjusted to a value that represents the training minimum. Learning on a mean-shift of $q$ of an arbitrary distribution over $x$ leads to $\sigma( w \cdot (x + q) + b )$, which increases the neuron response by $q$, weighted by $w$. During inference, both $w$ and $b$ are fixed, by extension the mean-shift $q$ is fixed as well. The mean-shift over larger inference data disappears, introducing an additional bias of $w \cdot q$ before non-linear activation. This training bias may lead to prediction errors of the neuron and consequently the full neural network.

\subsection{Windowed Aliasing}
\begin{figure}[h]
    \centering
    \includegraphics[width=\textwidth]{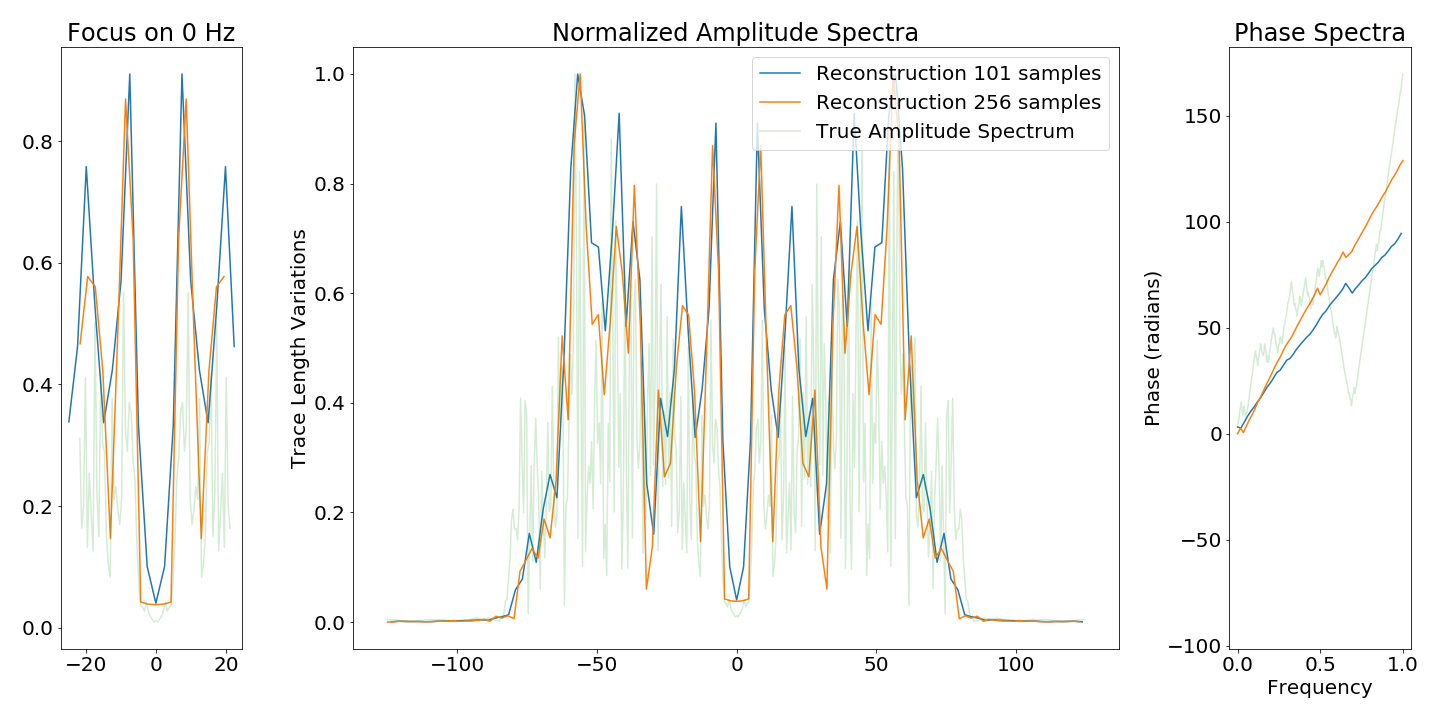}
    \caption{Spectral aliasing dependent on window-size (from \citet{Dramsch2018})}
    \label{fig:aliasing}
\end{figure}

Non-stationary data such as seismic data can contain sections within the data that contain spurious offsets from the mean. Figure~\ref{fig:aliasing} shows varying sizes of cutouts, with 101 and 256 samples respectively. In the middle, the full normalised amplitude spectra are presented. On the right, the corresponding phase spectra are presented. On the left, we focus on the frequency content of the amplitude spectra around 0 Hz. The cutouts were Hanning tapered, however, a mean shift appears with decreasing patch size.

These concepts of mean-shift corresponds to a DC offset in spectral data, which can be audio, seismic or electrical data. In images this corresponds to a non-zero alpha channel. While batch normalization can  correct the mean shift in individual mini-batches \citep{ioffe2015batch}, this may shift the entire spectrum by the aliased offset. Additionally, batch normalization may not be feasible in some physical applications pertaining to regression tasks.

\section{Complex Seismic Data}
Complex seismic traces are calculated by applying the Hilbert transform to the real-valued signal. The Hilbert transform applies a convolution with  to the signal, which is equivalent to a -90-degree phase rotation. It is essential that the signal does not contain a DC component, as this would not have a phase rotation.

The Hilbert transform is defined as 
\begin{equation}
    H(u)(t) = \frac{1}{\pi}\int_{-\infty}^\infty \frac{x(\tau)}{t-\tau}\,d\tau,
\end{equation}

of a real-valued time series $u(t)$, where the improper integral has to be interpreted as the Cauchy principal value. In the Fourier domain, the Hilbert transform has a convenient formulation, where frequencies are set zero and the remaining frequencies are multiplied by 2. This can be written as

\begin{equation}
    x_a = F^{-1}(F(x) 2U ) = x + iy
\end{equation}

where $x_a$ is the analytical signal, $x$ is the real signal, $F$ is the Fourier transform, and $U$ is the step function. The imaginary component $y$ is simultaneously the quadrature of the real-valued trace.
This provides locality to explicit phase information, where the Fourier transform itself does not lend itself to the resolution of the phase in the time domain. 
In conventional seismic trace analysis, the complex data is used to calculate the instantaneous amplitude and instantaneous frequency. These are beneficial seismic attributes for interpretation \citep{barnes2007tutorial}.

\section{Experiments}

\subsection{Data}
The data is the F3 seismic data, interpreted by \citet{alaudah2019machine}. They provide a seismic benchmark for machine learning with accessible NumPy format. The interpretation (labels) of the seismic data is relatively coarse compared to conventional seismic interpretation, but the accessibility and pre-defined test case are compelling.

The F3 data set was acquired in the Dutch North Sea in 1987 over an area of 375.31 km\textsuperscript{2}. The sampling-rates are 4~ms in time and inline/crossline bins of 25~m. The extent being 650 inline traces and 950 crossline traces with a total length of 1.848~s. The data contains faulted reflector packets, of which the lowest one overlays a salt diapir. The data contains some noise that masks lower-amplitude events.

We generate 64x64 2D patches in the inline and crossline direction from the 3D volume to train our network.
The fully convolutional architecture can predict on arbitrary sizes after training. 
The seismic data is normalized to values in the range of [-1, 1]. 
To obtain complex-valued seismic data we apply a Hilbert transform to every trace of the data and subtract the real-valued seismic from the real component \citep{taner1979complex}.

\subsection{Architecture}

\begin{table}
    \centering
    \begin{tabular}{|l|c|c|c|c|c|c|}
\toprule
Layer & \multicolumn{2}{c|}{Spatial} & $\mathbb{C}$omplex & $\mathbb{R}$eal & $\mathbb{C}$omplex & $\mathbb{R}$eal\\
(Size) & X & Y & Small & Small & Large & Large\\
\midrule 
Input & 64 &  64 & 2 & 1 &  2 &  1\\
\midrule
$(\mathbb{C})$-Conv2D & 64 &  64 &  8 & 8 &  16 &  16\\
\midrule
$(\mathbb{C})$-Conv2D + BN & 64 &  64 &  8 & 8 &  16 &  16\\
\midrule
Pool + $(\mathbb{C})$-Conv2D + BN & 32 &  32 &  16 & 16 & 32 & 32\\
\midrule
Pool + $(\mathbb{C})$-Conv2D + BN & 16 &  16 &  32 & 32 & 64 & 64\\
\midrule
Pool + $(\mathbb{C})$-Conv2D + BN & 8 &  8 &  64 & 64 & 128 & 128\\
\midrule
\midrule
Pool + $(\mathbb{C})$-Conv2D & 4 &  4 &  128 & 128 & 256 & 256\\
\midrule
\midrule
Up + $(\mathbb{C})$-Conv2D + BN & 8 &  8 & 64 & 64 & 128 & 128\\
\midrule
Up + $(\mathbb{C})$-Conv2D + BN & 16 &  16 & 32 & 32 & 64 & 64\\
\midrule
Up + $(\mathbb{C})$-Conv2D + BN & 32 &  32 & 16 &  16 & 32 &32\\
\midrule
Up + $(\mathbb{C})$-Conv2D & 64 &  64 &  8 & 8 & 16 & 16\\
\midrule
$(\mathbb{C})$-Conv2D + BN & 64 &  64 & 8  &  8 & 16 &  16\\
\midrule
$(\mathbb{C})$-Conv2D & 64 &  64 & 2 & 1 &  2 & 1\\
\midrule
\midrule
Parameters on Graph & \multicolumn{2}{c|}{} & 100,226 & 198,001 &  397,442 & 790,945\\
\midrule
Size on Disk [MB]& \multicolumn{2}{c|}{} &  1.4 & 2.5 & 4.8 & 9.2\\
\bottomrule
    \end{tabular}
    \caption{Layers used in the four auto-encoders and according parameter count on the computational graph and size on disc. Complex-valued convolutions and real-valued convolutions used respectively.}
    \label{tab:1}
    \vspace*{-1.1cm}
\end{table}

The Auto-encoder architecture uses 2D convolutions with 3x3 kernels. We employ batch normalization to regularize the training and speed up training \citep{ioffe2015batch}. 
The down and up sampling is achieved by MaxPooling and the UpSampling operation, respectively. 
We reduce a 64x64 input 4 times by a factor of two to encode a 4x4 encoding layer. 
The architecture for the complex convolutional network is identical, except for replacing the real-valued 2D convolutions with complex-valued convolutions. 
The layers used are shown below (see Table~\ref{tab:1}).

Complex-valued neural networks contain two feature maps for every feature map contained in a real-valued network. 
Matching real-valued and complex-valued neural networks is quite complicated, as the same filter values yield a vastly different amount of parameters, as can be seen in Table~\ref{tab:1}.
The smaller real-valued network contains as many feature maps for the real-valued seismic as the large complex network, the large complex network contains an additional feature map for every real-valued input for the complex component.
We define a complex-valued network that effectively has the same number of filters as the real-valued small network. This network effectively has half the available feature maps for the real-valued seismic input. 
Moreover, we define a large real-valued network to match the number of filters of the large complex-valued network, this network has twice the feature-maps available for representation of the real-valued seismic data, compared to the large complex-valued network. 
The parameters are counted on the computational graph compiled by Tensorflow.

\subsection{Training}
\begin{figure}[H]%
    \centering
    \includegraphics[width=.75\linewidth]{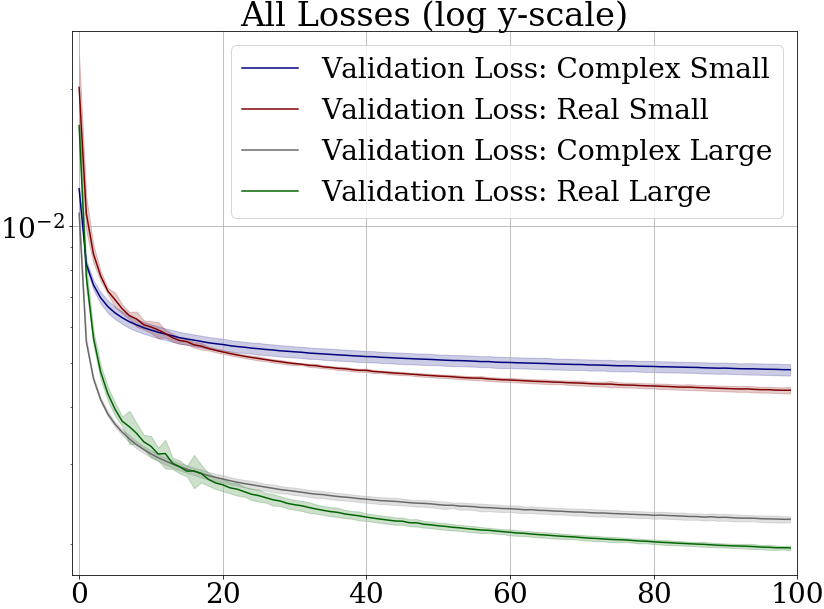}
    \caption{Validation Loss (MSE) on 7 random seeds per network. (Real-valued loss on real-valued seismic and combined complex-valued loss on complex-valued seismic, as the network "sees" it.)}%
    \label{fig:loss}%
\end{figure}

We train the networks with an Adam optimizer \citep{kingma2014adam} and a learning rate of $10^{-3}$ without decay, for 100 epochs. The loss function is mean squared error, as the seismic data contains values in the range of [-1,1]. All networks reach stable convergence without overfitting, shown in Figure~\ref{fig:loss}. 

\subsection{Evaluation}
We compare the complex auto-encoders with the real-valued auto-encoders, through the reconstruction error on unseen test data on 7 individual realizations of the respective four networks and qualitative analysis of reconstructed images. We focus on evaluating the real-valued reconstruction of the seismic data.
\section{Results}

We trained four neural network auto-encoders with seven random initializations for each network, to allow for error bars on the estimates in Figure~\ref{fig:loss}. The mean squared error and the mean absolute error for each parameter configuration during training is given in Table~\ref{tab:2}. There is a clear correspondence of the reconstruction error of the auto-encoder to the size of network. The real-valued networks outperform the complex-valued networks in both the mean squared error and mean absolute error, however, we see that a real-valued network needs around twice as many parameters as a complex-valued network to attain the same reconstruction errors.

\begin{figure}
    \centering
    \hspace*{-1cm}
    \includegraphics[width=1.2\textwidth]{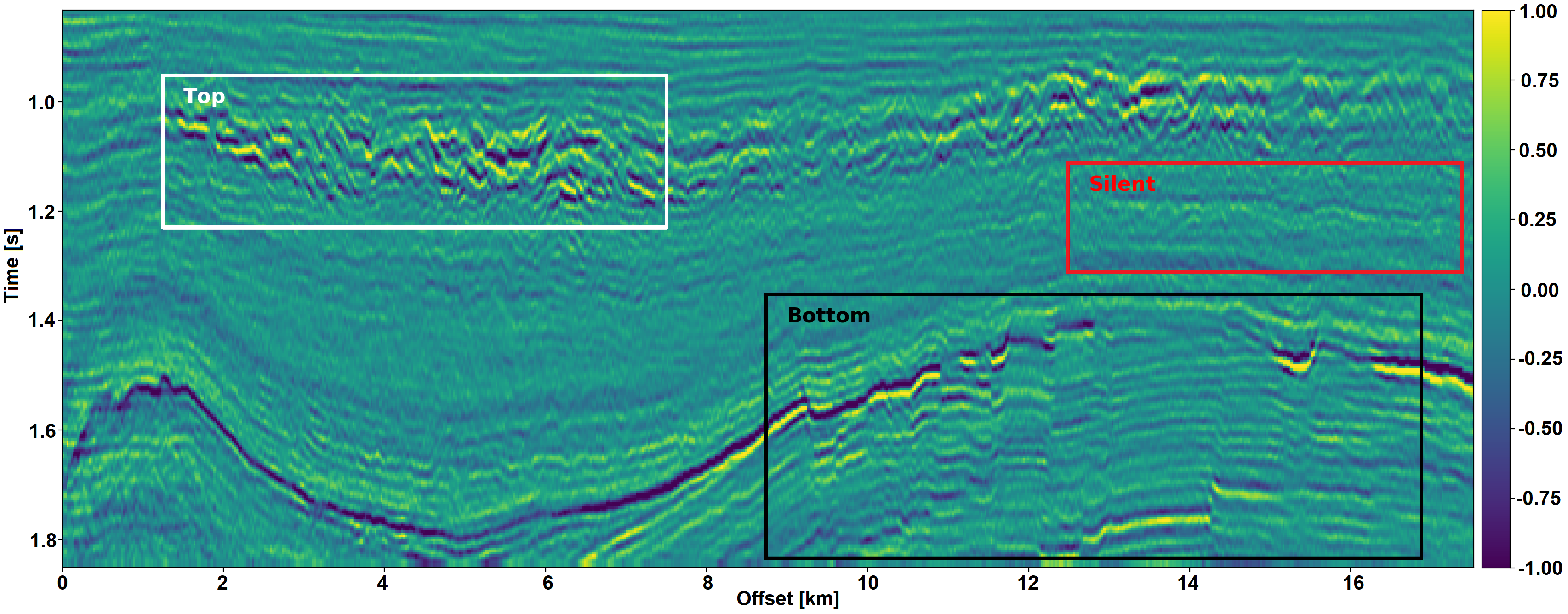}
    \caption{Seismic Test Data with marked section for closer inspection. We chose the "top" section for it's faulted chaotic texture, "bottom" for the faulted blocks, and "silent" for a noisy but geologically uninteresting section.}
    \label{fig:eval_seis}
\end{figure}

\begin{table}[H]
    \centering
    \begin{tabular}{|l|c|c|c|c|}
        \toprule
        Network & Runs $\times$ epochs & Parameters & MSE $[\times 10^{-2}]$ & MAE $[\times 10^{-2}]$\\
        \midrule
        1) $\mathbb{C}_\text{small}$ & $7\times100$ & 100,226 & 0.484 $\pm$ 0.013 & 4.695 $\pm$ 0.058 \\
        2) $\mathbb{R}_\text{small}$ & $7\times100$ & 198,001 & 0.436 $\pm$ 0.006 & 4.500 $\pm$ 0.028 \\
        3) $\mathbb{C}_\text{large}$ & $7\times100$ & 397,442 & 0.227 $\pm$ 0.003 & 3.247 $\pm$ 0.025 \\
        4) $\mathbb{R}_\text{large}$ & $7\times100$ & 790,945 & 0.196 $\pm$ 0.002 & 3.050 $\pm$ 0.013 \\
        \bottomrule
    \end{tabular}
    \caption{Parameters and errors for networks (lower is better). Losses on network validation.}
    \vspace*{-12pt}
    \label{tab:2}
\end{table}

\begin{figure}
    \centering
     \subfloat[.5\textwidth][Ground Truth]{\includegraphics[width=.5\linewidth]{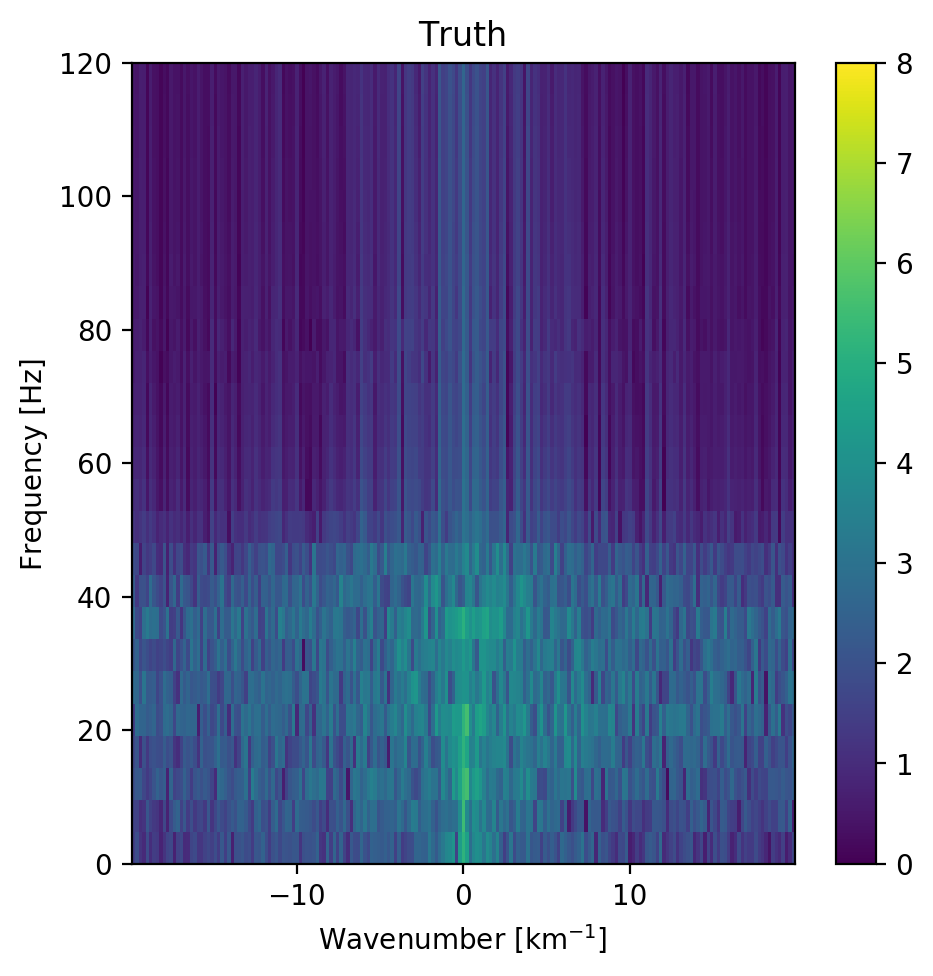}\label{fig:silent_truth_fk}}
     \subfloat[.5\textwidth][Large Complex Network]{\includegraphics[width=.5\linewidth]{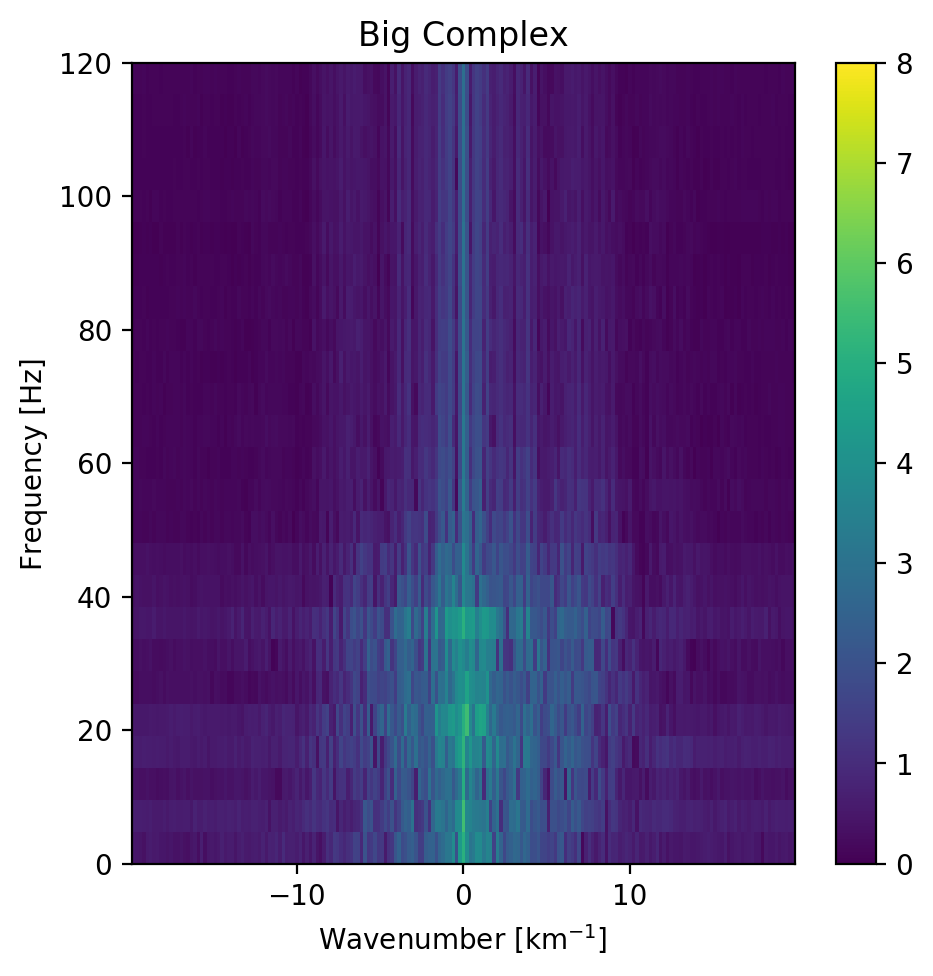}\label{fig:silent_lc_fk}}\\
     \caption{Evaluation on Silent Noise Patch in FK Domain. Noise reduction of frequencies below 50~Hz apparent, while reconstruction does not introduce visible aliasing.}
    \label{fig:silent_fk}
\end{figure}

The seismic sections in Figure~\ref{fig:eval_seis} show the unseen test seismic section. We perform a closer inspection of the regions "top" and "bottom" to focus on geologically relevant sections in the reconstruction process. The noisy segment without strong reflectors is a good baseline to evaluate the noise reduction of the Autoencoder and the behaviour of the different networks on low amplitude data. Overall, all networks denoise the original seismic, with the lowest reconstruction errors being RMS of 0.1187 and MAE of 0.0947 (cf. Table~\ref{tab:errors}). Figure~\ref{fig:silent_fk} shows the frequency-wavenumber (FK) of the ground truth (\ref{fig:silent_fk}~(a)) and the large complex network reconstruction (\ref{fig:silent_fk}~(b)). These show a decrease in the 0~-~60~Hz band for larger absolute wavenumbers.

\begin{table}[H]
    \centering
    \begin{tabular}{|l|c|c|c|c|c|c|c|c|}
        \toprule
                & \multicolumn{2}{c|}{Full}  & \multicolumn{2}{c|}{Silent} & \multicolumn{2}{c|}{Top} & \multicolumn{2}{c|}{Bottom}\\
        Network & RMS & MAE & RMS & MAE & RMS & MAE & RMS & MAE \\
        \midrule
        1) $\mathbb{C}_\text{small}$ & 0.1549  & 0.1145 & 0.1265  & 0.1010 & 0.2315  & 0.1759 & 0.1588  & 0.1200 \\
        2) $\mathbb{R}_\text{small}$ & 0.1581  & 0.1153 & 0.1247  & 0.0994 & 0.2395  & 0.1810 & 0.1612  & 0.1205 \\
        3) $\mathbb{C}_\text{large}$ & 0.1508  & 0.1101 & 0.1187  & 0.0947 & 0.2301  & 0.1747 & 0.1514  & 0.1135 \\
        4) $\mathbb{R}_\text{large}$ & 0.1469  & 0.1072 & 0.1214  & 0.0967 & 0.2222  & 0.1679 & 0.1459  & 0.1088 \\
        \bottomrule
    \end{tabular}
    \caption{RMS and MAE on real component of Data Patches.}
    \vspace*{-12pt}
    \label{tab:errors}
\end{table}

\subsection{"Top" seismic section}
\begin{figure}
    \centering
     \subfloat[\textwidth][Ground Truth]{\includegraphics[width=\linewidth]{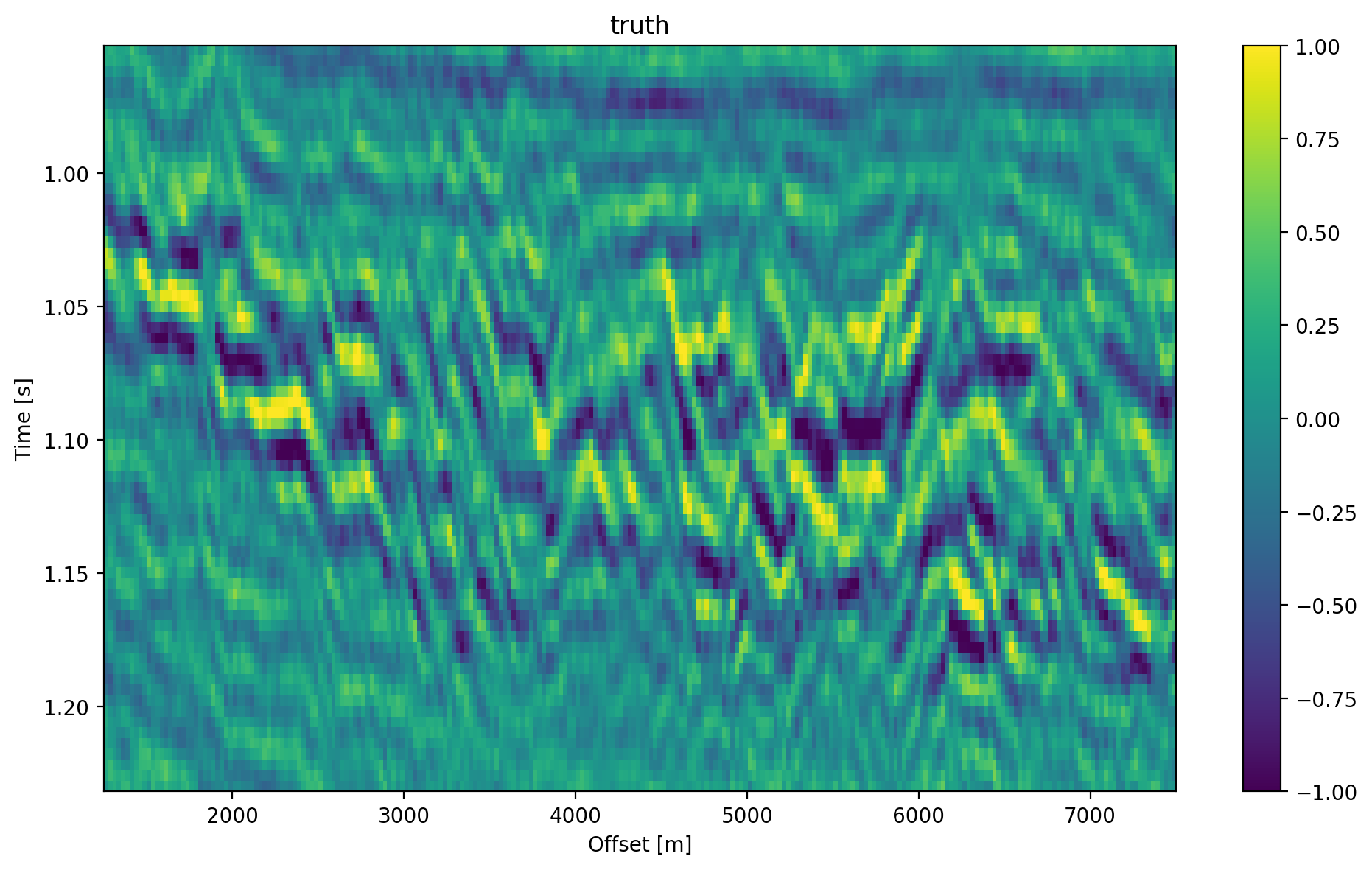}\label{fig:top_truth}}\\
     \subfloat[.5\textwidth][Small Complex Network top Patch]{\includegraphics[width=.5\linewidth]{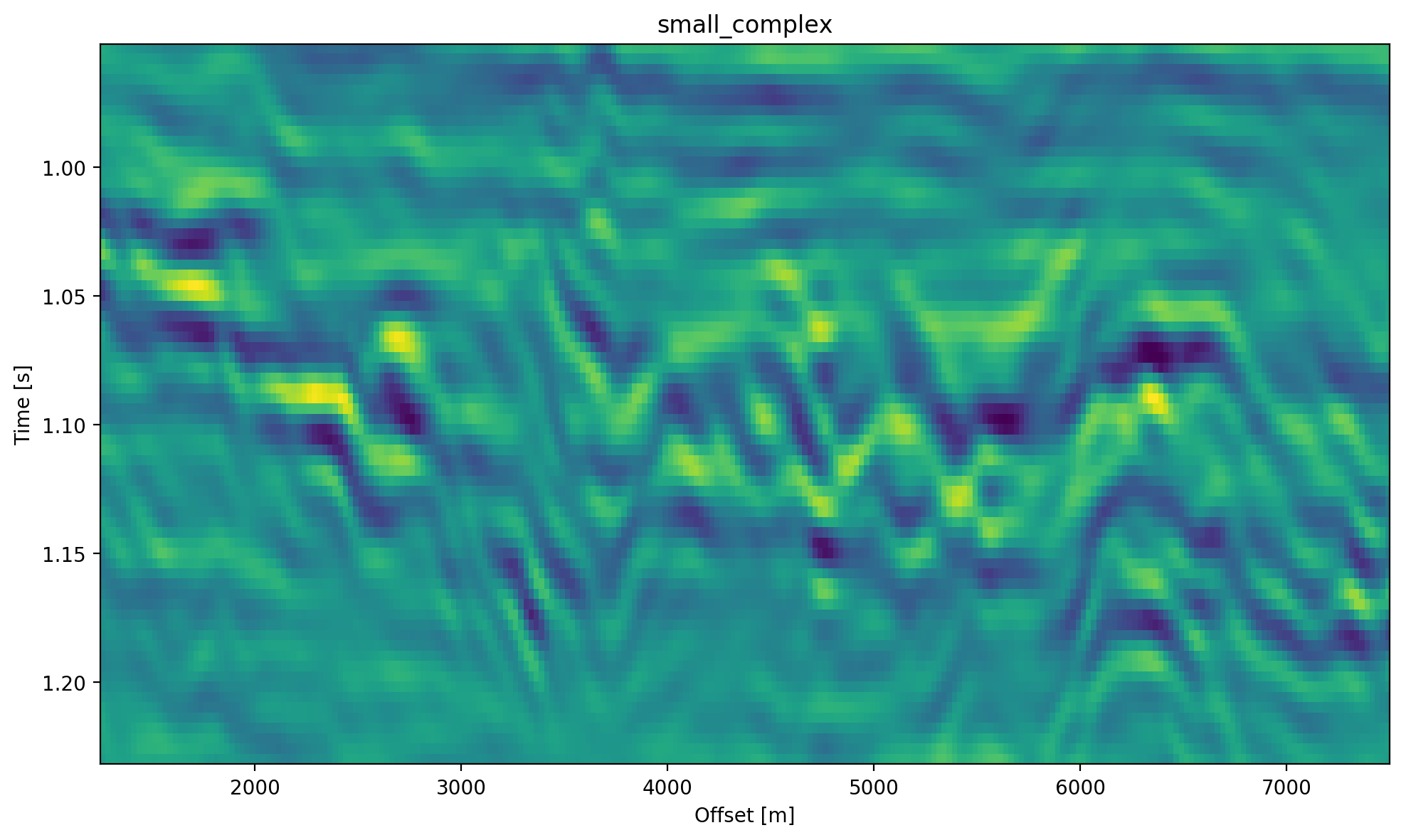}\label{fig:top_sc}}
     \subfloat[.5\textwidth][Small Real Network top Patch]{\includegraphics[width=.5\linewidth]{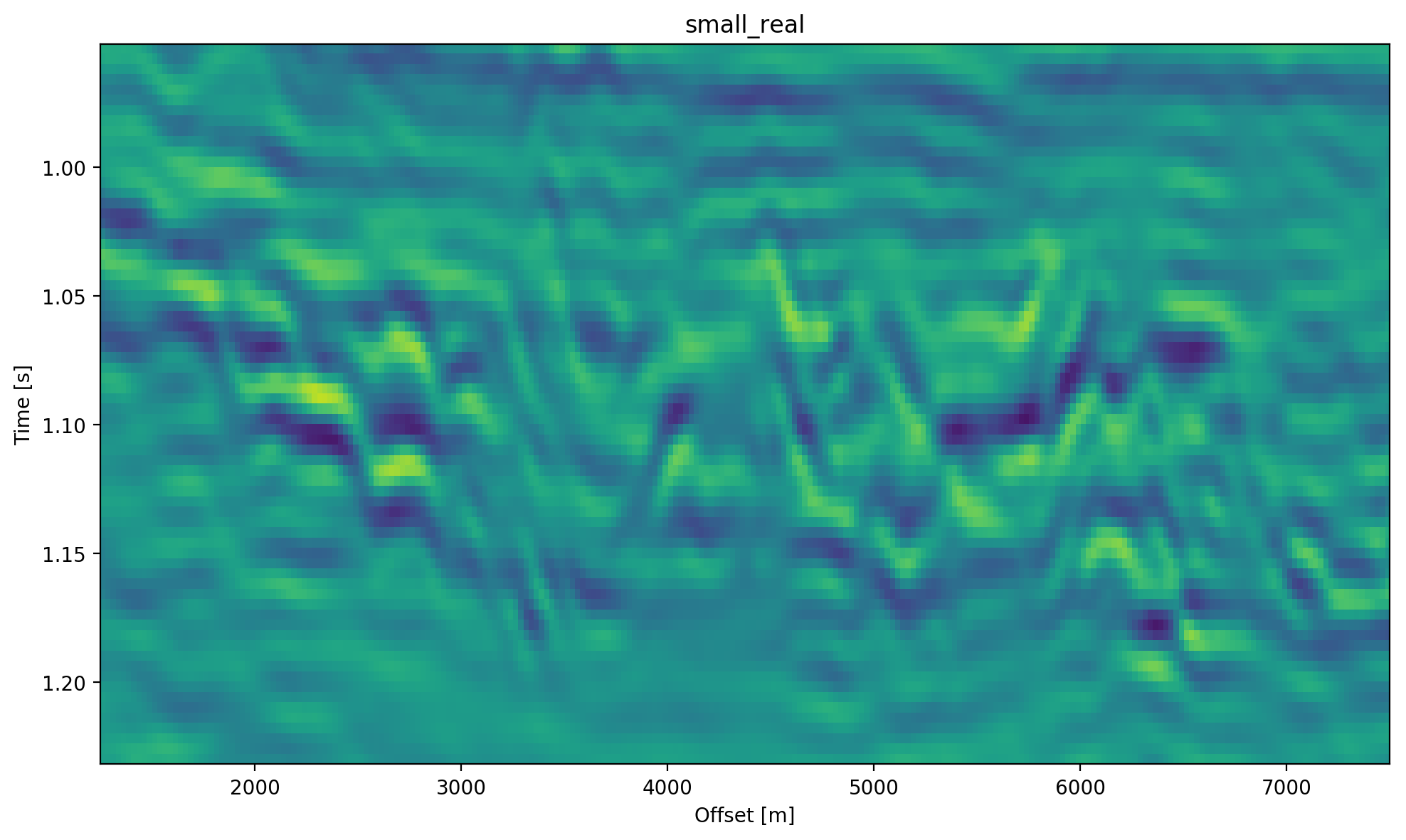}\label{fig:top_sr}}\\
     \subfloat[.5\textwidth][Large Complex Network top Patch]{\includegraphics[width=.5\linewidth]{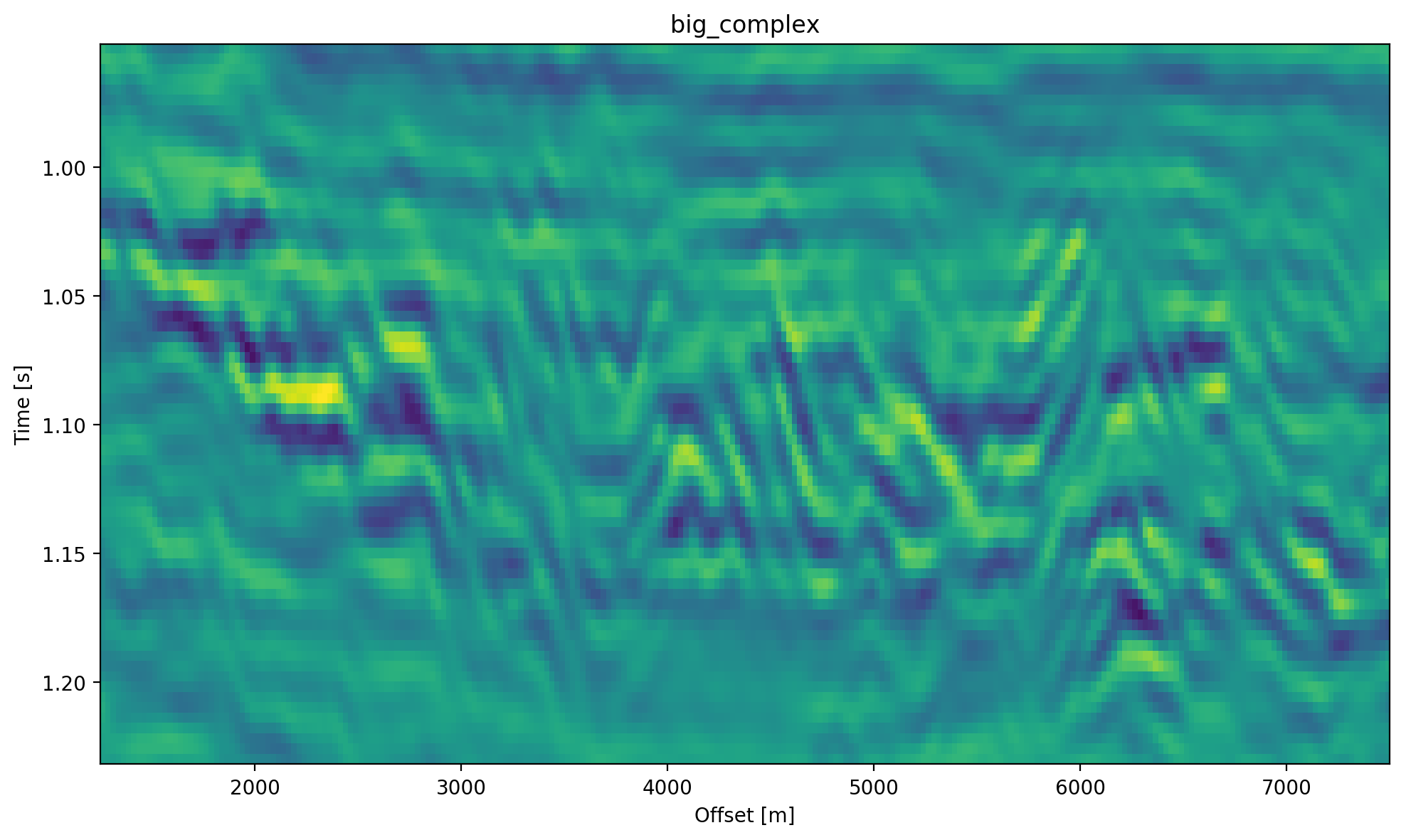}\label{fig:top_bc}}
     \subfloat[.5\textwidth][Large Real Network top Patch]{\includegraphics[width=.5\linewidth]{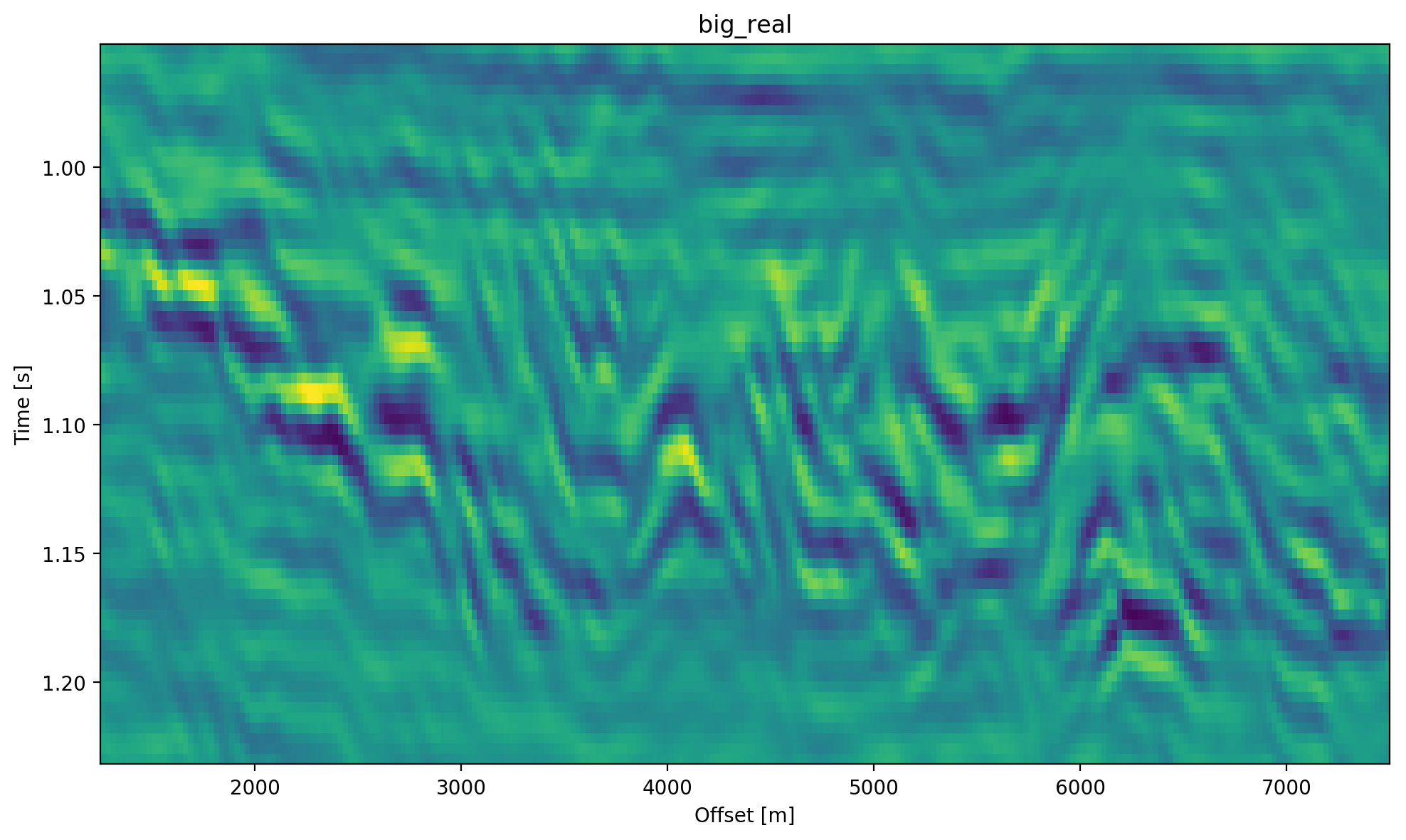}\label{fig:top_br}}\\
    \caption{Evaluation on top Noise Patch}
    \label{fig:top}
\end{figure}

The "top" segment contains strong reflections that are very faulted with strong reflectors. Figure~\ref{fig:top} shows the top segment and the reconstructions of the four networks. All networks display various amounts of smoothing. The quantitative results show that the complex networks perform very similar regardless of size. The large real-valued network outperforms the complex networks by 2.5~\% on RMS, while the small real-valued network underperforms by 2.5~\% on RMS. The panel in Figure~\ref{fig:top_sr} shows a very smooth result. Despite the close score of the complex networks, it appears that the complex-valued network restores more high-frequency content. We can also see less smearing of discontinuities in the larger complex network, particularly visible in the lower part (1.2~s) at 6000~m offset, which is smeared to appear like a diffraction in the smaller network. The large real-valued network shows good reconstruction with minor smearing with higher amplitude fidelity in areas like 1.1~s at 2000~m, however, some of the steeply dipping artifacts are visible below the reflector packet between 0~m and 2000~m offset.



\subsection{"Bottom" seismic section}

\begin{figure}
    \centering
     \subfloat[\textwidth][Ground Truth]{\includegraphics[width=\linewidth]{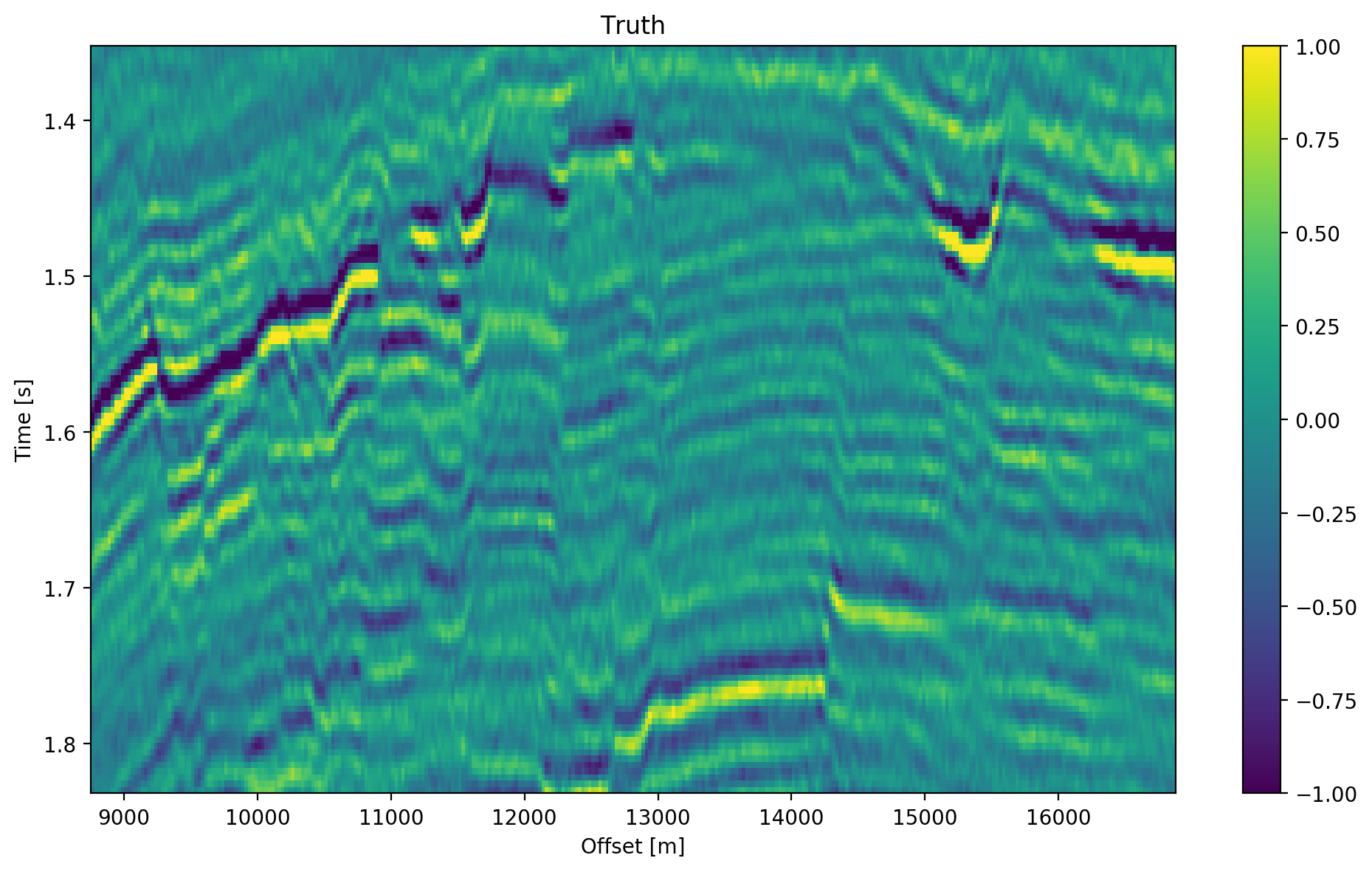}\label{fig:bottom_truth}}\\
     \subfloat[.5\textwidth][Small Complex Network bottom Patch]{\includegraphics[width=.5\linewidth]{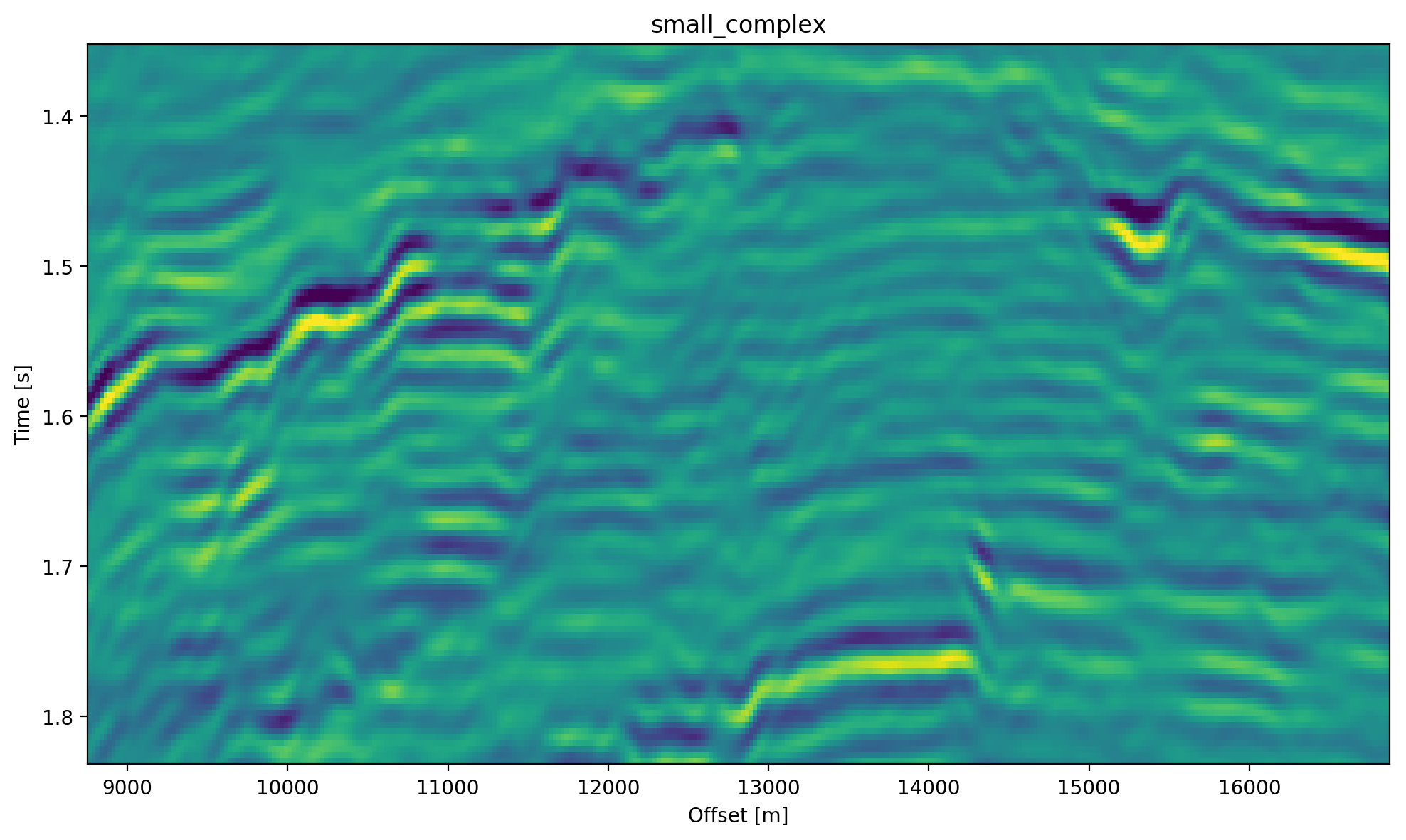}\label{fig:bottom_sc}}
     \subfloat[.5\textwidth][Small Real Network bottom Patch]{\includegraphics[width=.5\linewidth]{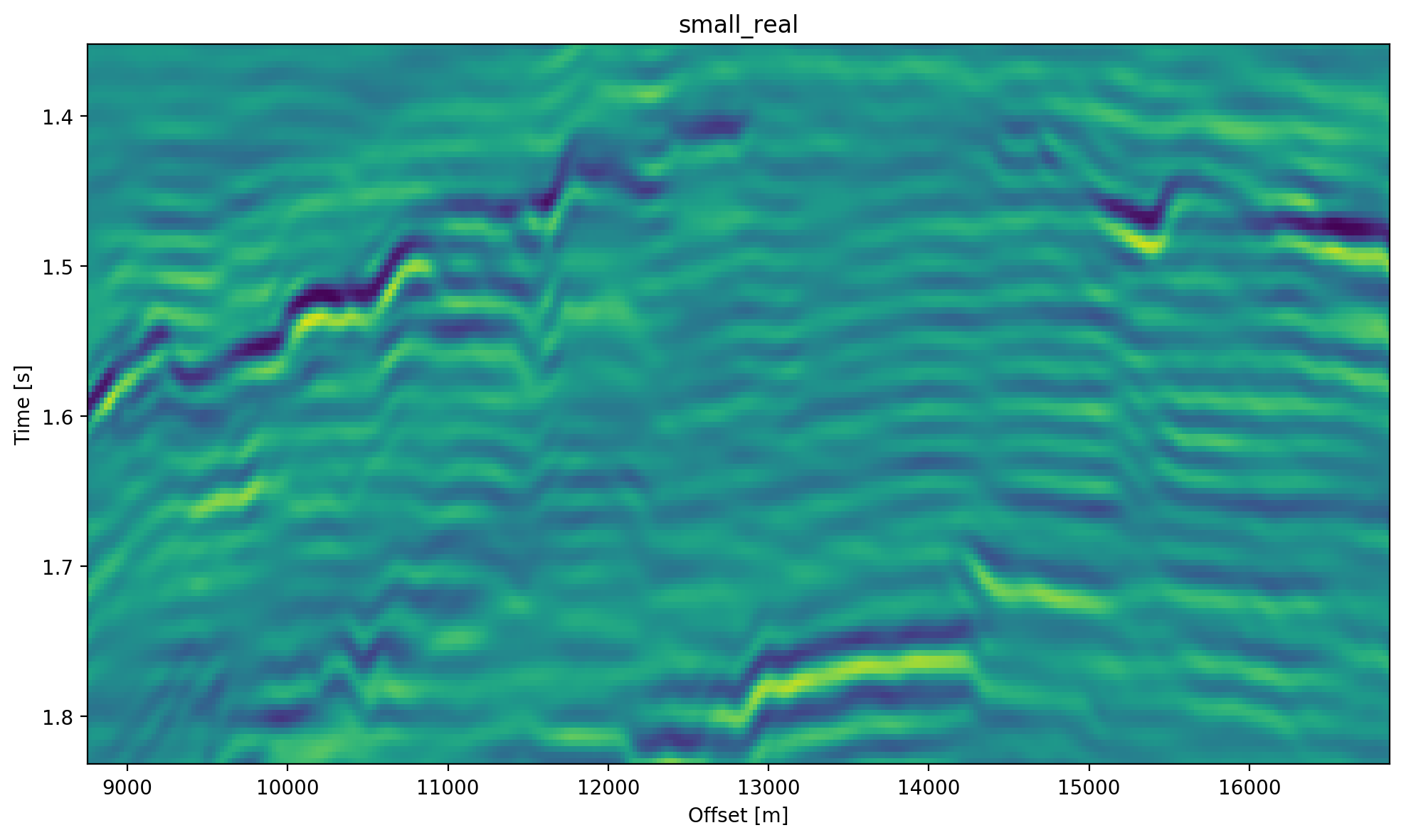}\label{fig:bottom_sr}}\\
     \subfloat[.5\textwidth][Large Complex Network bottom Patch]{\includegraphics[width=.5\linewidth]{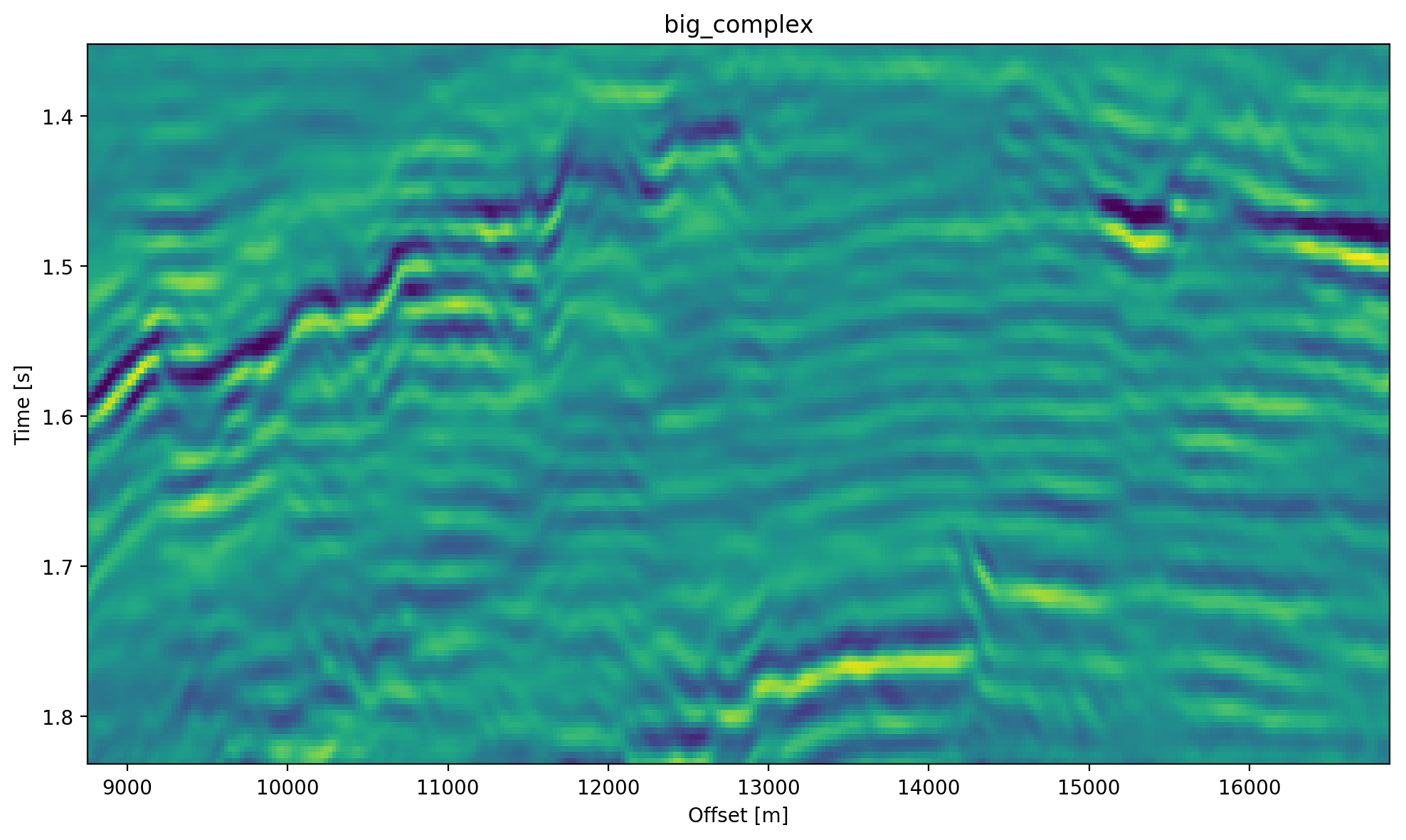}\label{fig:bottom_bc}}
     \subfloat[.5\textwidth][Large Real Network bottom Patch]{\includegraphics[width=.5\linewidth]{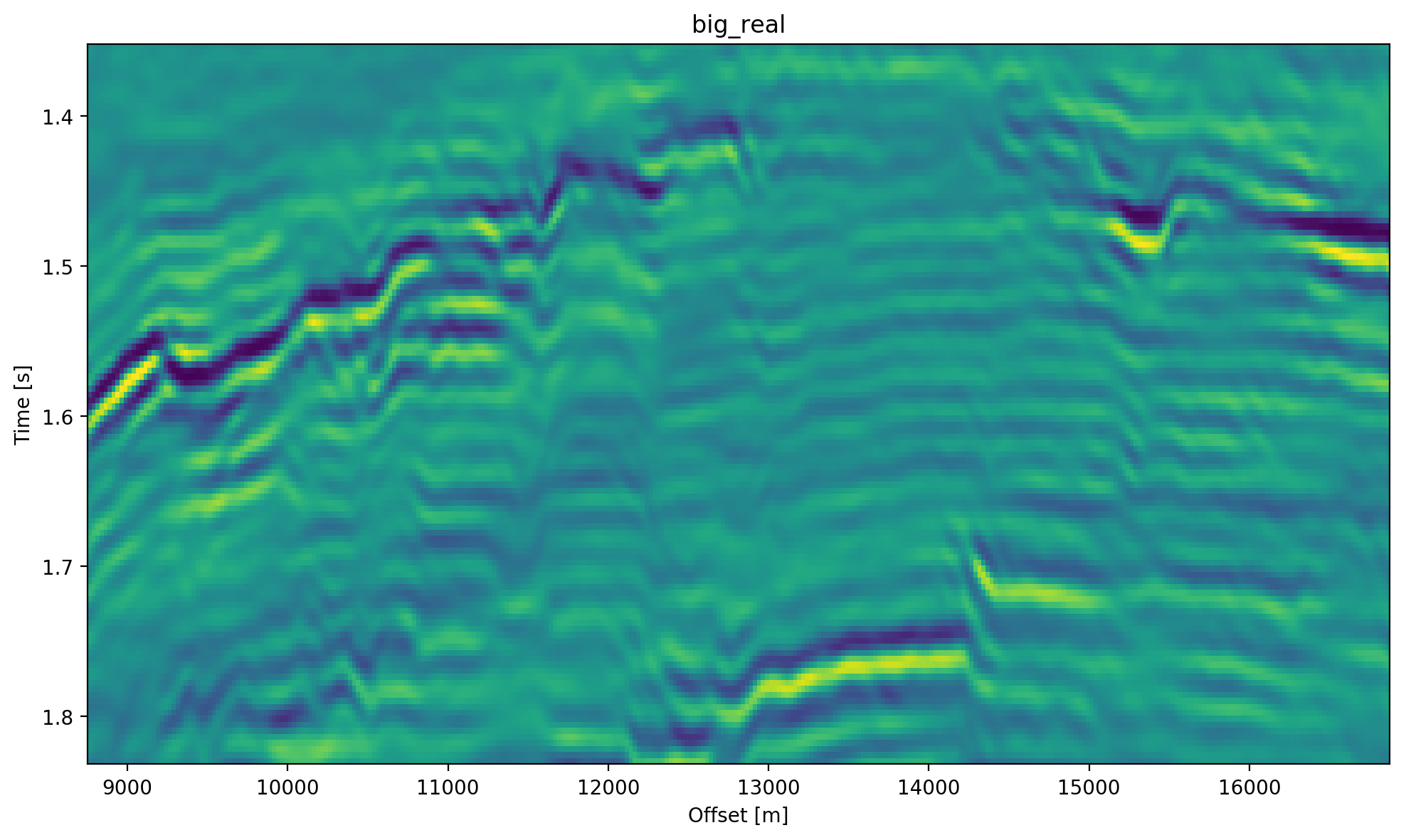}\label{fig:bottom_br}}\\
    \caption{Evaluation on bottom Noise Patch}
    \label{fig:bottom}
\end{figure}

The data marke as "bottom" in Figure~\ref{fig:eval_seis} contains a faulted anticline and relatively strong noise levels. The small complex network in Figure~\ref{fig:bottom_sc} reconstructs a denoised image with good reconstruction of the visible discontinuities. Some leakage of the reflector starting at 1.5~s across discontinuities is visible. The real small network in Figure~\ref{fig:bottom_sr} reconstructs a strongly smoothed image, with some ringing below the main reflector, which is not visible in the other reconstructions. The dipping reflector at an offset of 16000~m is well reconstructed, however, it seems like the reconstruction introduced ringing noise over the vertical image. The large real-valued network in Figure~\ref{fig:bottom_br} performs best quantitatively (cf. Table~\ref{tab:errors}). The complex-valued large network in Figure~\ref{fig:bottom_bc} does a fairly good job at reconstructing the image, similar to the large real-valued network. However, the amplitude reconstruction of high-amplitude events particularly in the main reflector around 1.5~s is showing. 

\subsection{Full seismic test data}
It is evident, that the small real-valued network does not match the performance of the smaller complex-valued network, even less so when compared to the large complex-valued network. We therefore compare the large networks on the full seismic data.

\begin{figure}
    \centering
     \subfloat[.35\textwidth][Ground Truth]{\includegraphics[width=.35\linewidth]{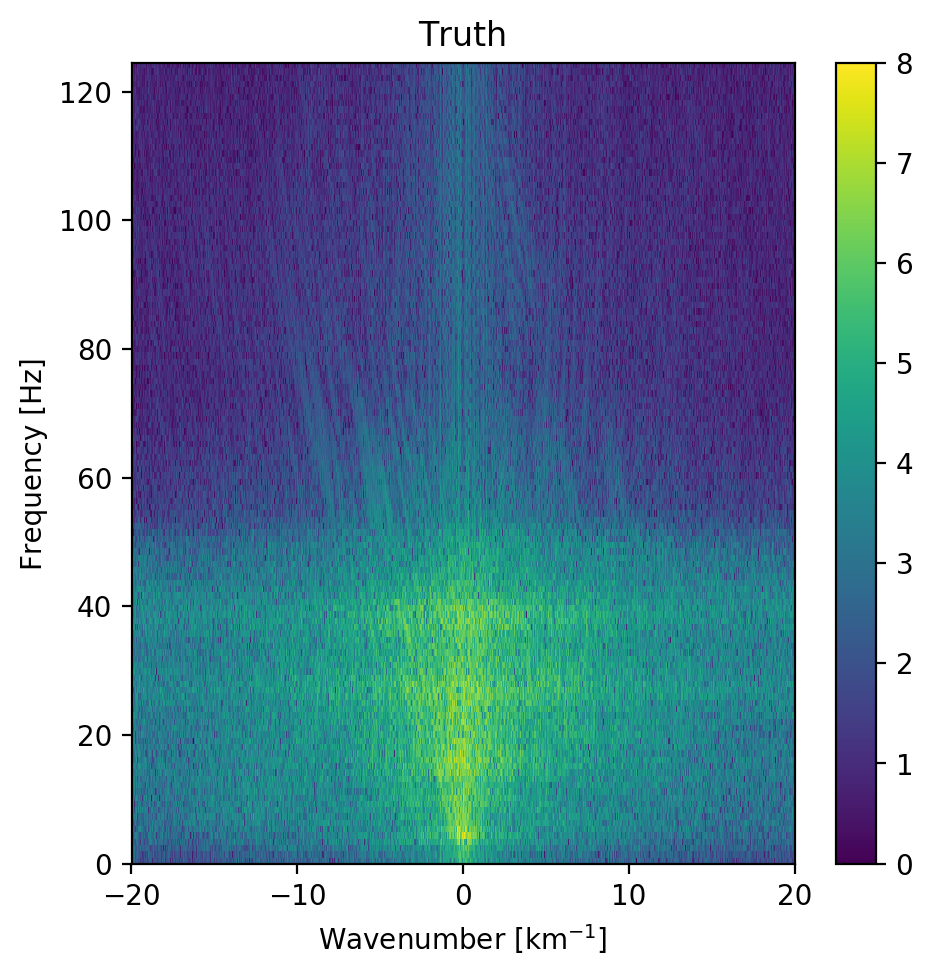}\label{fig:full_truth_fk}}
     \subfloat[.35\textwidth][Large Complex Network]{\includegraphics[width=.35\linewidth]{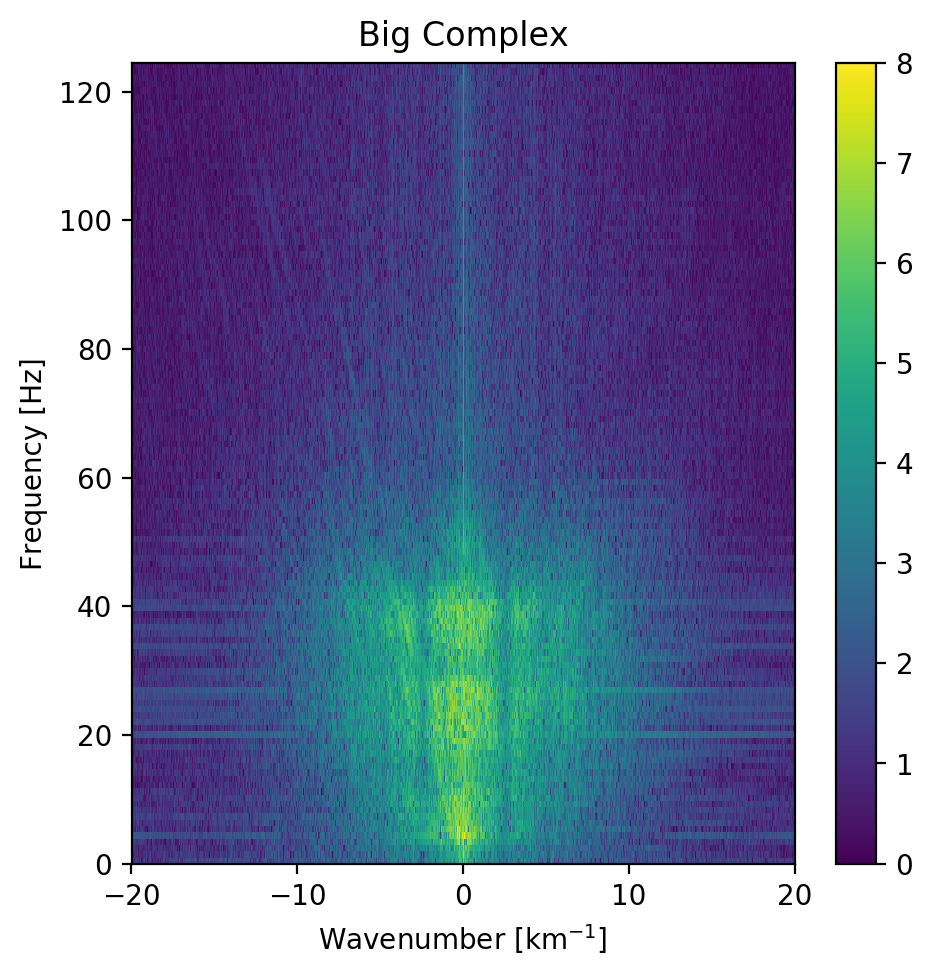}\label{fig:full_bc_fk}}
     \subfloat[.35\textwidth][Large Real Network]{\includegraphics[width=.35\linewidth]{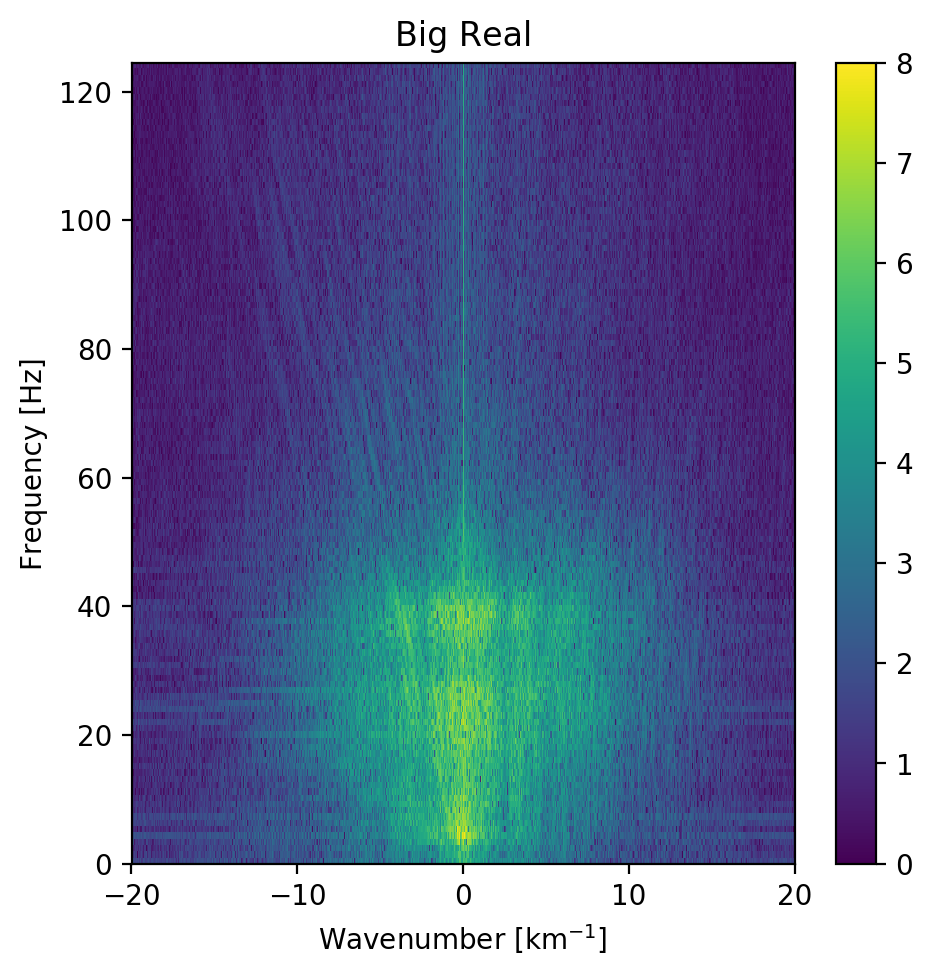}\label{fig:full_br_fk}}\\
    \caption{FK domain of full seismic data.}
    \label{fig:full_fk}
\end{figure}

\begin{figure}
    \centering
     \subfloat[\textwidth][Ground Truth]{\includegraphics[width=\linewidth]{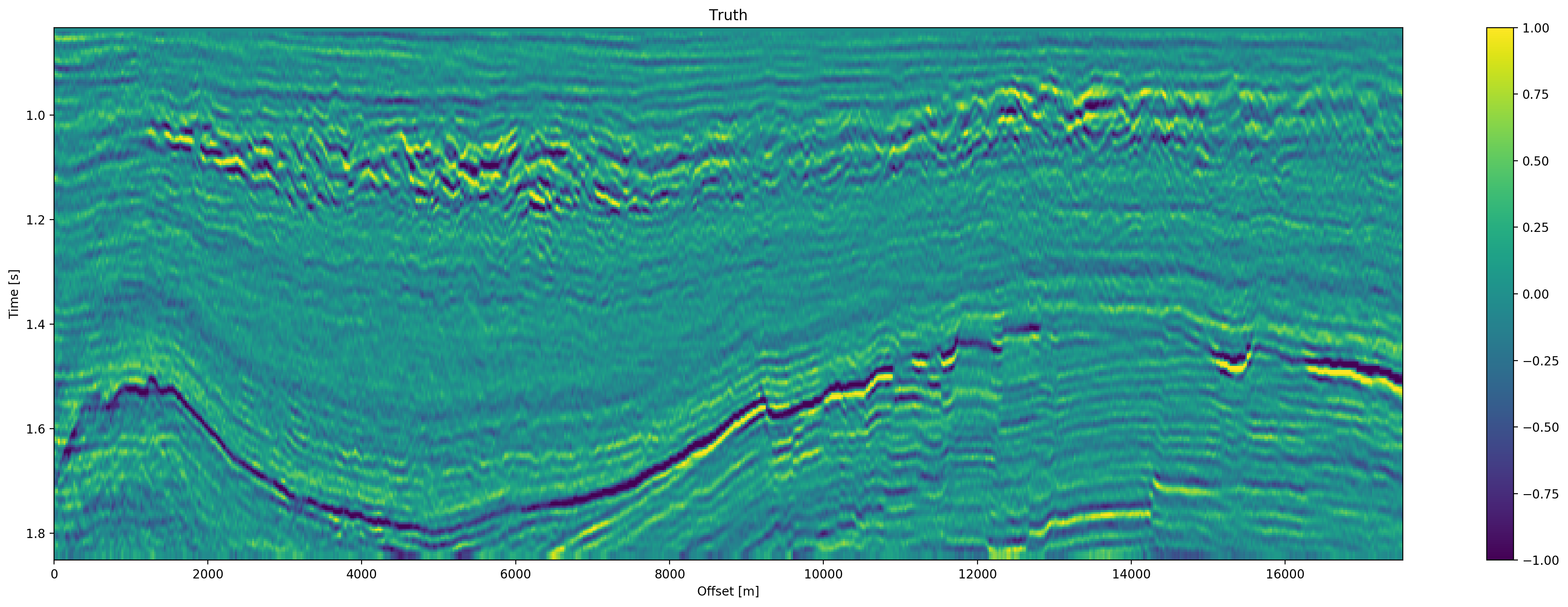}\label{fig:full_truth}}\\
     \subfloat[\textwidth][Large Complex Network]{\includegraphics[width=\linewidth]{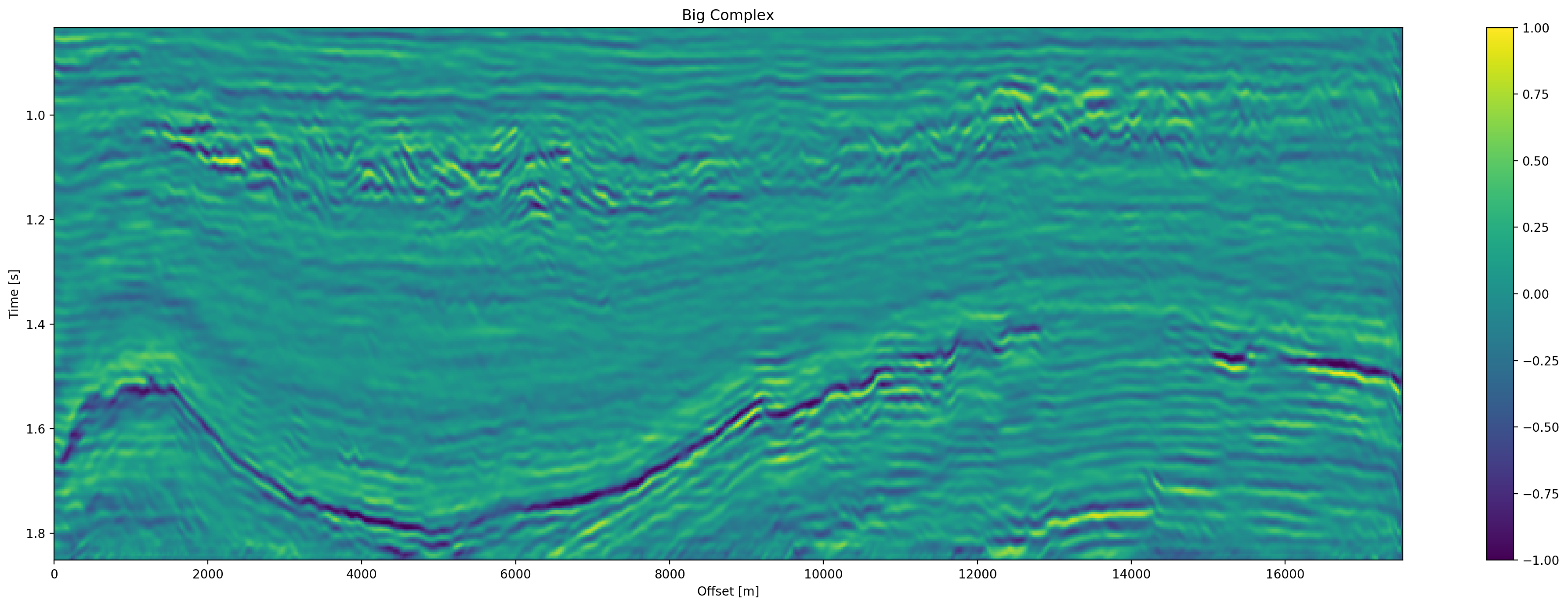}\label{fig:full_bc}}\\
     \subfloat[\textwidth][Large Real Network]{\includegraphics[width=\linewidth]{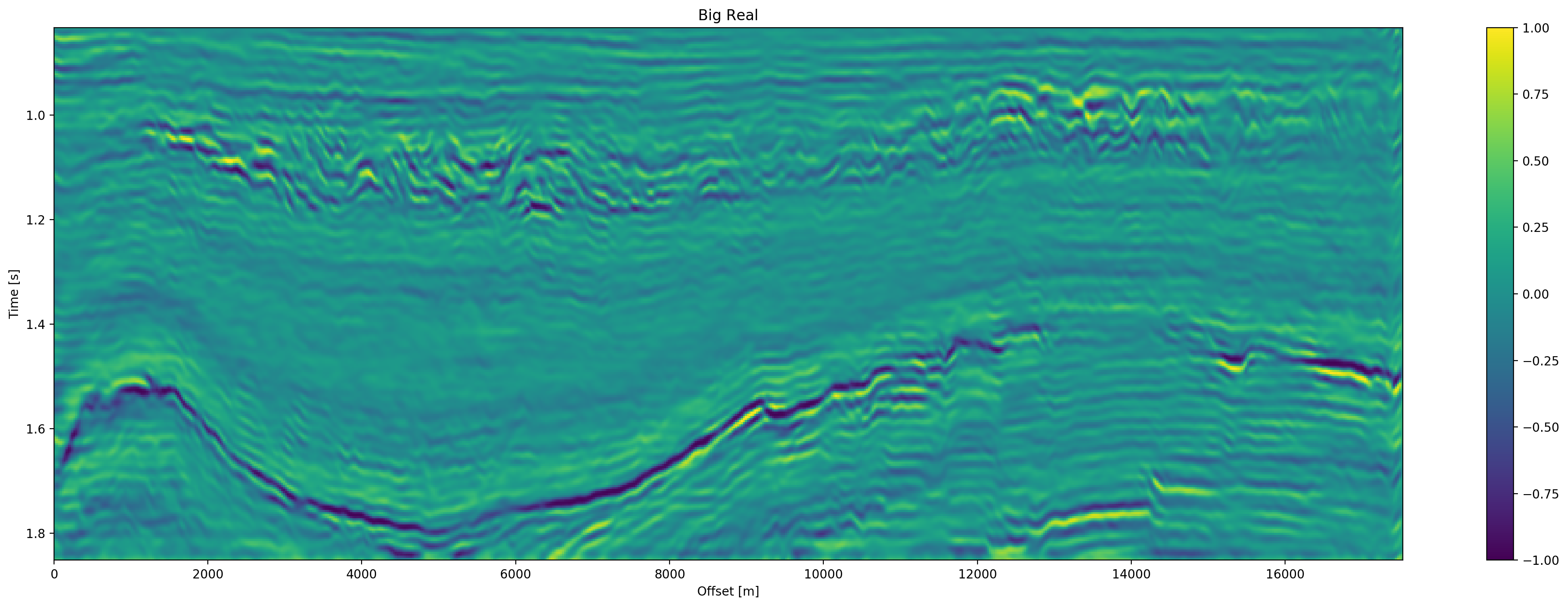}\label{fig:full_br}}
    \caption{Evaluation on full seismic data.}
    \label{fig:full}
\end{figure}

Overall, both networks return a smoothed image.
The findings for the strongly faulted sections in the "top" panel hold across the entire faulted area around 1.1~s in Figure~\ref{fig:full}.
The complex-valued network does a better job at reconstructing faults and discontinuities.
The real-valued network is better at reconstructing high-amplitude regions that appear dimmer in the complex-valued region.
The reconstruction of both networks seems adequately close to the ground truth, with differences in the details.
Quantitatively, the real-valued network does the better reconstruction in Table~\ref{tab:errors} with an improvement of 2.5~\% over the large complex-valued network.
The FK domain shows a very similar reduction in noise in the sub 50~Hz band in Figure~\ref{fig:full_fk}.
All networks introduce an increase of energy across all frequencies at wave-number $k=0~km^{-1}$.
Additionally, a dimming of the frequencies around $k=2.5~km^{-1}$ appears in all reconstructions, but is more prominent in the large complex-valued network.
The ground truth seismic contains some scattered energy in the high-frequency mid-wavenumber region, visible as "diagonal stripes". These were attenuated in the complex-valued network in Figure~\ref{fig:full_bc_fk}, but are partially present in the real-valued reconstruction in Figure~\ref{fig:full_br_fk}.


\section{Discussion}
We evaluated the outputs of the real-valued and complex-valued neural networks. All auto-encoder outputs are blurred to different degrees and denoised. The denoising effect of the seismic was most visible in the frequency band below 50~Hz. Additionally, some scattered high-frequency energy was attenuated by the networks.

The largest differences of the outputs in real-valued and complex-valued networks can be observed in discontinuous areas. Particularly, the faulted blocks in the top quarter and in the bottom center of the seismic section show inconsistencies. The real-valued network smooths over discontinuities and steep reflectors. Fault lines are imaged better in the complex-valued network output. 

In seismic data processing, including phase information stabilizes discontinuities and disambiguates cycle-skipping in horizons. This could be observed in the network performance and reconstruction. The increase in performance of the real-valued networks was significant (7.0~\% RMS), while the complex-valued networks already had an acceptable performance on the smaller network architecture  (2.6~\% RMS). We provide the complex-valued networks with a bias towards learning phase information, by providing the Hilbert transformed analytical trace, while the real-valued network needs to learn this information implicitly from the data itself. Considering, that during the training, the complex network evaluates both the real-valued seismic, which we primarily care about in addition to the complex-valued component, we can see how the losses in Figure~\ref{fig:loss} differ from the real-valued networks.

The largest network with 790,945 trainable parameters quantitatively performed the best on the reconstruction of the data. However, analysis of the reconstructed seismic shows, that while the high-amplitude regions are reconstructed to higher fidelity, discontinuous sections may be smeared by the real-valued network. The real-valued network that was matched to contain as many filters for the real-valued component of the seismic as the large complex-valued network, did not perform well. Furthermore, the smaller complex-valued network with 100,226 parameters that contains as many filter maps as the real-valued network in total, and half the trainable parameters, outperformed the smaller real-valued network across all test cases.
\section{Conclusion}
The inclusion of phase-information leads to a better representation of seismic data in convolutional neural networks. Complex-valued networks perform consistently, where real-valued networks have to learn phase-representations through implicit correlation, which requires larger networks. We show that complex trace information in deep neural networks improves the imaging of discontinuities as well as steep reflectors, particularly in chaotic seismic textures that are smoothed by real-valued neural networks of the same size. 

We show that convolutional neural networks can perform lossy compression on seismic data, where the reconstruction error is dependent on both network architecture and implementation details, like providing explicit phase information. During this compression, noise and scattered energy get attenuated. The real-valued network is prone to introduce steeply dipping artifacts in the reconstruction.

The stabilization of the reconstruction can be useful in seismic applications. While automatic seismic interpretation may benefit from the inclusion of information on discontinuities, we see the main application to be lossy seismic compression. The open source tool developed to make this research possible, enables further research and development of complex-valued solutions to non-stationary physics problems that benefit from explicit phase information.

The research shows that a change as small as 2.5~\% in RMS can change the reconstruction from being acceptable to very smeared to a geoscientist. This touches on the fact that better metrics to evaluate computer vision tasks in geoscience are necessary. Additionally, these tasks have to be noise-robust and while amplitude-preserving be outlier robust too. Moreover, more research in the frequency dimming of bands in the network reconstruction is necessary.

Overall, the computational memory footprint of the complex convolution is higher than real-valued convolutional neural networks comparing singular convolutional operations. A significant increase in depth and width of networks to obtain an acceptable result in real-valued neural network to implicitly learn the phase information is necessary. The complex-valued networks an 8\textsuperscript{th} of the size already performs well, suggesting that expert domains that contain beneficial information in the phase of signals, could benefit from applying complex convolutional networks.
\section{Acknowledgments}
We thank Andrew Ferlitsch for his valuable insights. The research leading to these results has received funding from the Danish Hydrocarbon Research and Technology Centre under the Advanced Water Flooding program. We thank DTU Compute for access to the GPU Cluster.




\small
\bibliographystyle{plainnat}  
\bibliography{mybibfile}


\end{document}